\documentclass[10pt,journal,compsoc]{IEEEtran}
\usepackage{graphicx}
\usepackage{marvosym}
\usepackage{amsmath}
\usepackage{autobreak}
\usepackage{amssymb}
\usepackage{multirow}
\usepackage{diagbox}
\usepackage{tablefootnote}
\usepackage{threeparttable}
\usepackage{soul}
\usepackage[noend]{algpseudocode}
\usepackage{algorithmicx,algorithm}
\usepackage[colorlinks]{hyperref}
\usepackage{color}   
\definecolor{blush}{rgb}{0.62,0.24,0.81}
\definecolor{royalazure}{rgb}{0.0, 0.22, 0.66}
\definecolor{uared}{rgb}{0.85, 0.0, 0.3}

\definecolor{black}{rgb}{0,0,0}
\newcommand{\rone}[1]{\textcolor{black}{#1}}
\newcommand{\rtwo}[1]{\textcolor{black}{#1}}
\newcommand{\ronetwo}[1]{\textcolor{black}{#1}}

\ifCLASSOPTIONcompsoc
  \usepackage[nocompress]{cite}
\else
  \usepackage{cite}
\fi
\ifCLASSINFOpdf
\else
\fi
\begin{document}
%
\title{Category-Level 6D Object Pose Estimation with Flexible Vector-Based Rotation Representation}
%
%
%
%


\author{Wei~Chen{\Letter}, Xi~Jia, Zhongqun~Zhang, Hyung~Jin~Chang, Linlin~Shen, Jinming~Duan and ~Ale\v{s}~Leonardis
	       
\IEEEcompsocitemizethanks
{\IEEEcompsocthanksitem 
wei.chen.ai@outlook.com
}

}

\markboth{Journal of \LaTeX\ Class Files,~Vol.~x, No.~x, xxxx~20xx}%
{Shell \MakeLowercase{\textit{et al.}}: Bare Demo of IEEEtran.cls for Computer Society Journals}
%



\IEEEtitleabstractindextext{
\begin{abstract}
In this paper, we propose a novel 3D graph convolution based pipeline for category-level 6D pose and size estimation from monocular RGB-D images. The proposed method leverages an efficient 3D data augmentation and a novel vector-based decoupled rotation representation. 
Specifically, we first design an orientation-aware autoencoder with 3D graph convolution for latent feature learning. 
The learned latent feature is insensitive to point shift and size thanks to the shift and scale-invariance properties of the 3D graph convolution. 
Then, to efficiently decode the rotation information from the latent feature, we design a novel flexible vector-based decomposable rotation representation that employs two decoders to complementarily access the rotation information. 
The proposed rotation representation has two major advantages: 1) decoupled characteristic that makes the rotation estimation easier; 2) flexible length and rotated angle of the vectors allow us to find a more suitable vector representation for specific pose estimation task.
Finally, we propose a 3D deformation mechanism to increase the generalization ability of the pipeline.
Extensive experiments show that the proposed pipeline achieves state-of-the-art performance on category-level tasks. Further, the experiments demonstrate that the proposed rotation representation is more suitable for the pose estimation tasks than other rotation representations.
\end{abstract}

}

\maketitle

\IEEEdisplaynontitleabstractindextext

%
\IEEEpeerreviewmaketitle

\IEEEraisesectionheading{\section{Introduction}\label{sec:introduction}}

%
%
%
%
%
%
%
%
%
%

\IEEEPARstart{E}{stimating} 3D orientation and 3D translation, i.e., 6D pose, of rigid objects plays an essential role in many computer vision tasks such as augmented reality \cite{marchand2016pose}, virtual reality \cite{burdea2003virtual}, and smart robotic arm \cite{zhu2014single,tremblay2018deep}. 
For instance-level 6D pose estimation, in which training set and test set contain the same objects, substantial progress has been made \cite{Xiang2017,Rad2017bb8,Oberweger2018,Li_2018_ECCV,he2020pvn3d}. 
However, category-level 6D pose estimation is still challenging as the object shape and color are various in the same category.
Existing methods addressed this problem by mapping the different objects in the same category into a uniform model/map via RGB feature or RGB-D fusion feature. 
For example, Wang \textit{et al.} \cite{wang2019nocs} trained a modified Mask R-CNN \cite{he2017mask} to predict the normalized object coordinate space (NOCS) map of different objects based on RGB feature, and then computed the pose with observed depth and NOCS map by Umeyama algorithm \cite{umeyama1991least}. 
Chen \textit{et al.} \cite{chen2020cass} proposed to learn a canonical shape space (CASS) to tackle intra-class shape variations with RGB-D fusion feature \cite{wang2019densefusion}.
Tian \textit{et al.} \cite{tian2020shapeprior} trained a network to predict the NOCS map of different objects, with the uniform shape prior learned from a shape collection in NOCS-Synthetic dataset \cite{wang2019nocs}, and RGB-D fusion feature \cite{wang2019densefusion}.

\begin{figure}[!t]
\centering
\includegraphics[width=1.0\linewidth]{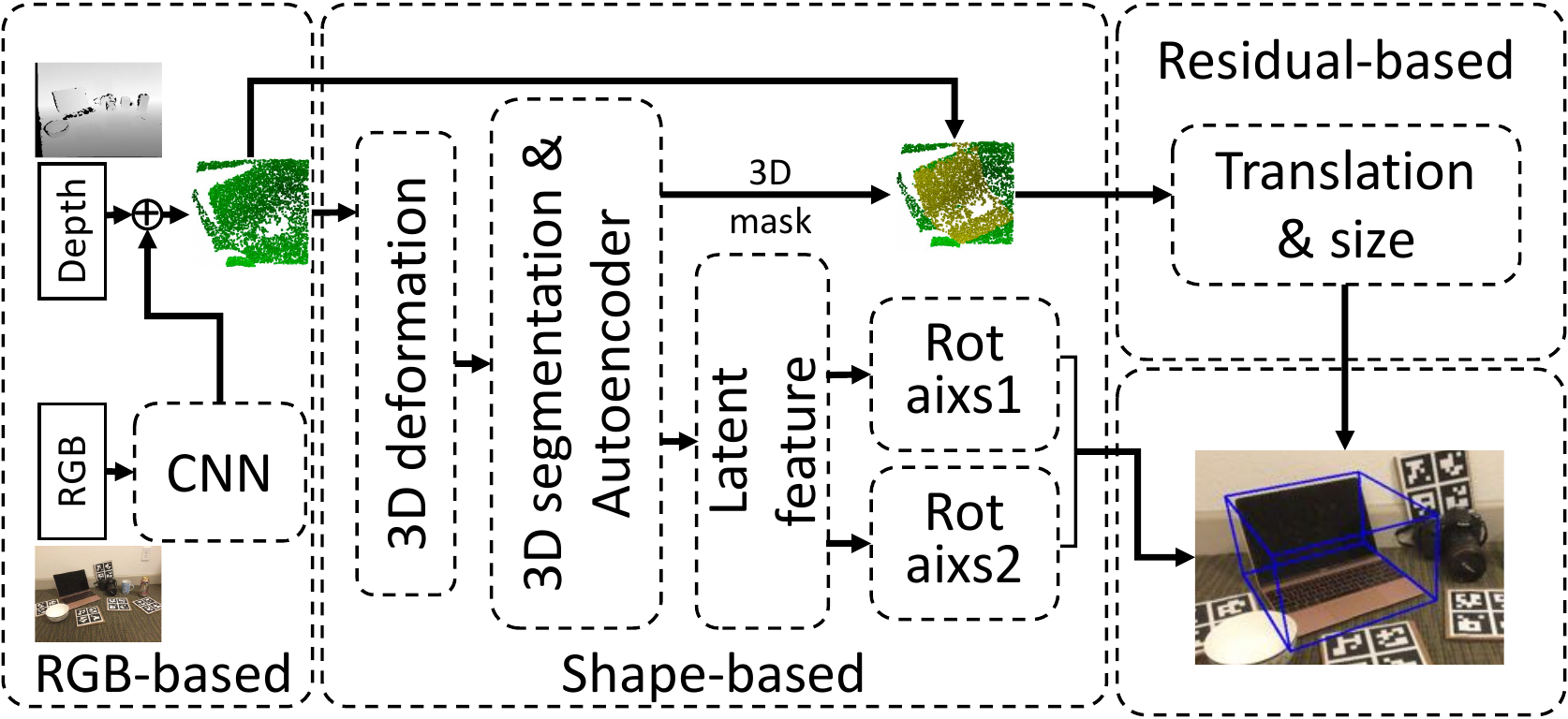}
   \caption{\textbf{Semantic illustration of FS-Net.} We use different networks for different tasks. The RGB-based network is used for 2D object detection, and the shape-based 3D graph convolution autoencoder is used for 3D segmentation and rotation estimation. The residual-based network is used for translation and size estimation with segmented points.}
\label{fig:short}
\vspace{-10pt}
\end{figure} 

Although these methods achieved state-of-the-art performance, there are still two issues.
Firstly, the benefits of using RGB feature or RGB-D fusion feature for category-level pose estimation are questionable. In \cite{vlach2016we}, Vlach et al. showed that people focus more on shape than color when categorizing objects, as different objects in the same category have very different colors but stable shapes (shown in Fig. \ref{fig:color_var}). Thereby, the use of RGB feature for category-level pose estimation can lead to low performance due to huge color variation in the test scene.
To address this issue, the previous method employed a large synthetic dataset \cite{wang2019nocs} to increase the generalization performance of the method. In contrast, to alleviate the color variation, we merely use the RGB feature for 2D detection, while using the shape feature learned with point cloud extracted from depth image for category-level pose estimation.

Secondly, learning a representative uniform shape requires a large amount of training data; therefore, the performance of these methods is not guaranteed with limited training examples. 
To overcome this issue, we propose a 3D graph convolution (3DGC) based autoencoder \cite{lin2020convolution} to effectively learn the category-level pose feature via observed points reconstruction of different objects instead of uniform shape mapping. 
We further propose an online 3D data augmentation mechanism for data augmentation to reduce the dependencies of labeled data.


Overall, the proposed framework, named FS-Net, consists of three parts: 2D detection, 3D segmentation \& rotation estimation, and translation \& size estimation. In the 2D detection part, we use the YOLOv3 \cite{redmon2018yolov3} to detect the object bounding box for coarse object points obtainment \cite{Chen_2020_CVPR}. 
Then in the 3D segmentation \& rotation estimation part, we design a 3DGC-based autoencoder to perform segmentation and observed points reconstruction jointly. 
The autoencoder encodes the orientation information in the latent feature. Then we design the flexible vector-based rotation (FVR) representation that uses two decoders to decode the category-level rotation information. Further, the flexible lengths and orientation of the vectors make it possible to find a more suitable representation for specific task.
For translation and size estimation, since they are all point coordinates related, we design a coordinate residual estimation network based on PointNet \cite{Qi_2017_CVPR} to estimate the translation residual and size residuals. 

To further increase the generalization ability of FS-Net, we use the proposed online 3D deformation for data augmentation.
%
To summarise, the main contributions of this paper are as follows:
\begin{itemize}

    \item We propose a 3DGC-based autoencoder to reconstruct the observed points for latent orientation feature learning and estimate category-level 6D object size and pose. Due to the efficient category-level pose feature extraction, the framework runs at 20 FPS on a GTX 1080Ti GPU.

    \item {To fully decode the rotation information from the learned feature in pose estimation tasks, we design a new flexible FVR representation and we provide a grid-search based solution to find the optimal parameters for FVR representation.}
    
    \item Based on the shape similarity of intra-class objects, we propose a novel 3D deformation mechanism to augment the training data.

      
\end{itemize}

\begin{figure*}[htp!]
\begin{center}
\includegraphics[width=0.999\linewidth]{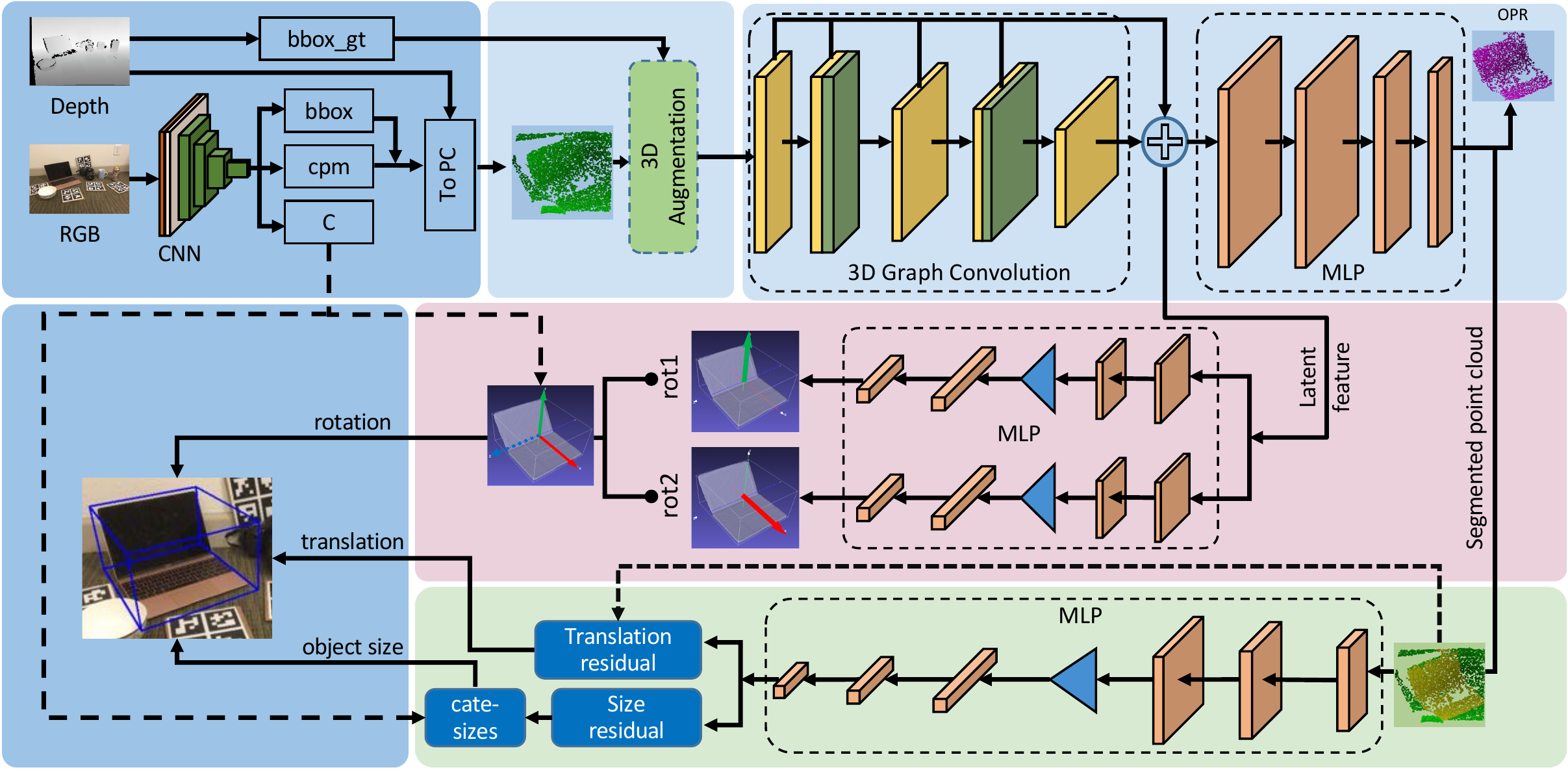}

\end{center}
\caption{\textbf{Architecture of FS-Net.} The input of FS-Net is an RGB-D image. For RGB channels, we use a CNN-based 2D detector to detect the object 2D location, category label `C' (used for next tasks), and class probability map (cpm) (generate the 3D sphere center via maximum probability location and camera parameters). With this information and depth channel, the point cloud (PC) in a compact 3D sphere are generated. Given the PC in the 3D sphere, we first use the proposed 3D augmentation mechanism for data augmentation. After that, we use a shape-based 3DGC autoencoder to perform observed points reconstruction (OPR), as well as point cloud segmentation, for orientation latent feature learning. Then we decode the rotation information into two perpendicular vectors from the latent feature. Finally, we use a residual estimation network to predict the translation and size residuals. `cate-sizes' denotes the pre-calculated average sizes of different categories, `$k$' is the rotation vector dimension, and the hollow `+' means feature concatenation. Please note, the 3D augmentation mechanism and bbox\_gt are only deployed during training.}
\label{fig:arch_whole}
\end{figure*}

\section{Related Works}
\subsection{Instance-Level Pose Estimation}
In instance-level pose estimation, a known 3D object model is usually available for training and testing. Based on the 3D model, instance-level pose estimation can be roughly divided into three types: template matching-based, correspondences-based, and voting-based methods. Template matching methods \cite{hinterstoisser2012gradient, Rad2018, Oberweger2018} aligned the template to the observed image or depth map via hand-crafted or deep learning feature descriptors. As they need the 3D object model to generate the template pool, their applications in category-level 6D pose estimation are limited. Correspondences-based methods trained their model to establish 2D-3D correspondences \cite{Rad2017bb8,Rad2018,peng2018pvnet, Li_2019_ICCV} or 3D-3D correspondences \cite{Chen_2020_CVPR, Chen_2020_WACV}. Then they solved perspective-n-point \cite{Gao2003} with 2D-3D or SVD problem with 3D-3D correspondences \cite{kabsch1976solution}. Some methods \cite{Chen_2020_WACV, Brachmann2016} also used these correspondences to generate voting candidates, and then used the RANSAC \cite{fischler1981random} algorithm for selecting the best candidate. However, the generation of canonical 3D keypoints is based on the known 3D object model that is not available when predicting the category-level pose.

\subsection{Category-Level Pose Estimation}
Compared to instance-level, one major challenge of category-level pose estimation is the intra-class object variation, including shape and color variation. To handle the object variation problem, \cite{wang2019nocs} proposed to map the different objects in the same category to a NOCS map. Then they used semantic segmentation to access the observed points cloud with known camera parameters. 
The 6D pose and size are calculated by the Umeyama algorithm \cite{umeyama1991least} with the NOCS map and the observed points. Shape-Prior \cite{tian2020shapeprior} adopted a similar method with \cite{wang2019nocs}, but both extra shape prior knowledge and dense-fusion feature \cite{wang2019densefusion}, instead of RGB feature, were used. 
CASS \cite{chen2020cass} estimated the 6D pose via the learning of a canonical shape space with dense-fusion feature \cite{wang2019densefusion}. Since the RGB feature is sensitive to color variation, the performance of their methods in category-level pose estimation is limited. In contrast, our method is based on shape feature which is robust for this task. 


\subsection{Data Augmentation}
In 3D object detection tasks \cite{Chen_2020_CVPR,Qi_2018_CVPR,shi2020pv, Chen_2020_WACV}, online data augmentation techniques such as translation, random flipping, shifting, scaling, and rotation are applied to original point clouds for training data augmentation. 
However, simply applying these operations on point clouds is unable to handle the shape variation problem in the 3D task as these operations cannot change the shape property of the object. To address this, \cite{choi2020part} proposed part-aware augmentation which operates on the semantic parts of the 3D object with five manipulations: dropout, swap, mix, sparing, and noise injection. However, how to decide the semantic parts are ambiguous. In contrast, we propose a box-cage-based 3D data augmentation mechanism that generates the various shape variants (shown in Fig. \ref{fig:3d_defor}) and avoids semantic parts decision procedure.

\subsection{Rotation Representation}
Widely used rotation representations, such as Euler angle representation, quaternion representation, and axis-angle representation, suffer from discontinuous issues \cite{zhou2019continuity}, which are not suitable for network learning. Although the $R_{6D}$ representation proposed in \cite{zhou2019continuity} is free from the discontinuous issue, they took two columns/rows from the $3\times3$ rotation matrix as their new representation. \ronetwo{
Derived from rotation matrix, R6D is essentially a fixed representation in the camera coordinate.
In contrast, our new FVR representation is defined based on two flexible orthogonal vectors and the vectors can change the lengths and mutual rotated angle to generate more suitable vector representations (see Fig. \ref{fig:vdr_6d}).}

\section{Proposed Method}
In this section, we describe the detailed architecture of FS-Net shown in Fig. \ref{fig:arch_whole}. 
Firstly, we use YOLOv3 \cite{redmon2018yolov3} to detect the object location with RGB input. Secondly, we use 3DGC \cite{Lin_2020_CVPR} autoencoder to perform 3D segmentation and observed points reconstruction, the latent feature can learn orientation information through the process.
Then we propose the novel FVR representation for decoding orientation information. Thirdly, we use PointNet \cite{Qi_2017_CVPR} to estimate the translation and object size. 
Finally, to increase the generalization ability of FS-Net and save storage space, we propose the box-cage based 3D deformation for data augmentation.

\begin{figure}[t!]
\begin{center}
\includegraphics[width=0.8\linewidth]{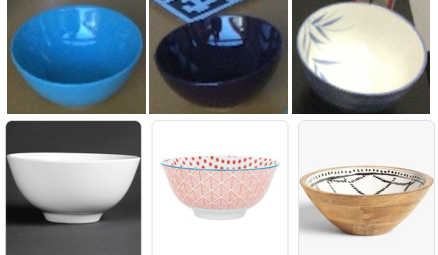}
\end{center}
   \caption{\textbf{Stable shape and various color.} Top row: three bowl instances randomly chosen from the NOCS-REAL dataset. Bottom row: three bowl instances randomly cropped from the internet image search results (using the keyword `bowl'). The color is varied, while the shape is relatively stable.}
\label{fig:color_var}
\vspace{-5pt}
\end{figure}

\subsection{Object Detection}
\label{sec:obj_det}

Following our previous work\cite{Chen_2020_CVPR}, we train a YOLOv3 \cite{redmon2018yolov3} to fast detect the object bounding box in RGB images, and output class (category) labels. Then we adopt the 3D sphere to locate the point cloud of the target object quickly. 
The 2D detection part provides a compact 3D learning space for the following tasks. 
Different from other category-level 6D object pose estimation methods \cite{wang2019nocs,chen2020cass,tian2020shapeprior} that need semantic segmentation masks, we only need object bounding boxes. Since object detection is faster and easier than semantic segmentation \cite{redmon2018yolov3, he2017mask, redmon2017yolo9000}, the detection speed of our method is faster than previous category-level pose estimation methods.

\subsection{Shape-Based Network}


The output points from object detection contain both object and background points. To access the points that belong to the target object and calculate the rotation of the object, we need a network that performs two tasks: 3D segmentation and rotation estimation. 

Although there are many network architectures that directly process point cloud \cite{Qi_2017_CVPR,qi2017pointnetplusplus,zhou2018voxelnet}, most of the architectures calculate on point coordinates, which means their networks are sensitive to point cloud shift and size variation \cite{lin2020convolution}. This decreases the pose estimation accuracy.

To tackle the point clouds shift, Frustum-PointNet \cite{Qi_2018_CVPR} and G2L-Net \cite{Chen_2020_CVPR} employed the estimated translation to align the segmented point clouds to local coordinate space. However, their methods cannot handle the intra-class size variation.

To solve the point clouds shift and size variation problem, in this paper, we propose a 3DGC autoencoder to extract the point cloud shape feature for segmentation and rotation estimation. 3DGC is designed for point cloud classification and object part segmentation; our work shows that 3DGC can also be used for category-level 6D pose estimation tasks.

\subsubsection{3D Graph Convolution}
3DGC kernel consists of $m$ unit vectors. The $m$ kernel vectors are applied to the $n$ vectors generated by the center point with its $n$-nearest neighbors. Then, the convolution value is the sum of the cosine similarity between $m$ kernel vectors and the $n$-nearest vectors. In a 2D convolution network, the trained network learned a weighted kernel, which has a high response with a matched RGB value, while the 3DGC network learned the orientations of the $m$ vectors in the kernel. The weighted 3DGC kernel has a high response with a matched 3D pattern that is defined by the center point with its $n$-nearest neighbors. For more details, please refer to \cite{lin2020convolution}.

\subsubsection{Rotation-Aware Autoencoder}
Based on the 3DGC, we design a 3DGC autoencoder for the estimation of category-level object rotation. To extract the latent rotation feature, we train the autoencoder to reconstruct the observed points transformed from the observed depth map of the object. There are several advantages to this strategy: 1) the reconstruction of observed points is view-based and symmetry invariant, 2) the reconstruction of observed points is easier than that of a complete object model, and 3) more representative latent feature can be learned (shown in Table \ref{tab:abla}).

In \cite{sundermeyer2020multi,sundermeyer2018implicit}, the authors reconstructed the input images to observed views as well. However, the input and output of their models are 2D images that are different from our 3D point cloud input and output. Furthermore, our network architecture is also different from theirs. Our work shows that this strategy can also be applied to point cloud reconstruction with the new architecture.

We utilize Chamfer Distance to train the autoencoder, the reconstruction loss function $\mathcal{L}_{rec}$ is defined as
\begin{equation}
\mathcal{L}_{rec} =
\sum_{x_i \in M_{c}} \min _{\hat{x}_i \in \hat{M}_{c}}\|x_i-\hat{x}_i\|_{2}^{2}
 +\sum_{\hat{x}_i \in \hat{M}_{c}} \min _{x_i \in M_{c}}\|x_i-\hat{x}_i\|_{2}^{2},
\end{equation}
where $M_c$ and $\hat{M}_c$ denote the ground truth point cloud and reconstructed point cloud, respectively. $x_i$ and $\hat{x}_i$ are the points in $M_c$ and $\hat{M}_c$. With the help of 3D segmentation mask, we only use the features extracted from the observed object points for reconstruction.

After the network convergence, the encoder learned the rotation-aware latent feature. 
Since the 3DGC is scale and shift-invariant, the observed points reconstruction enforces the autoencoder to learn the scale and shift-invariant orientation feature under corresponding rotation.
In the next subsection, we describe how we effectively decode rotation information from the learned latent feature.

\subsection{Flexible Vector-Based Rotation Representation}
\label{sec:rotde}

%

\begin{figure}[htp!]
\begin{center}
\includegraphics[width=0.8\linewidth,height=3.5cm]{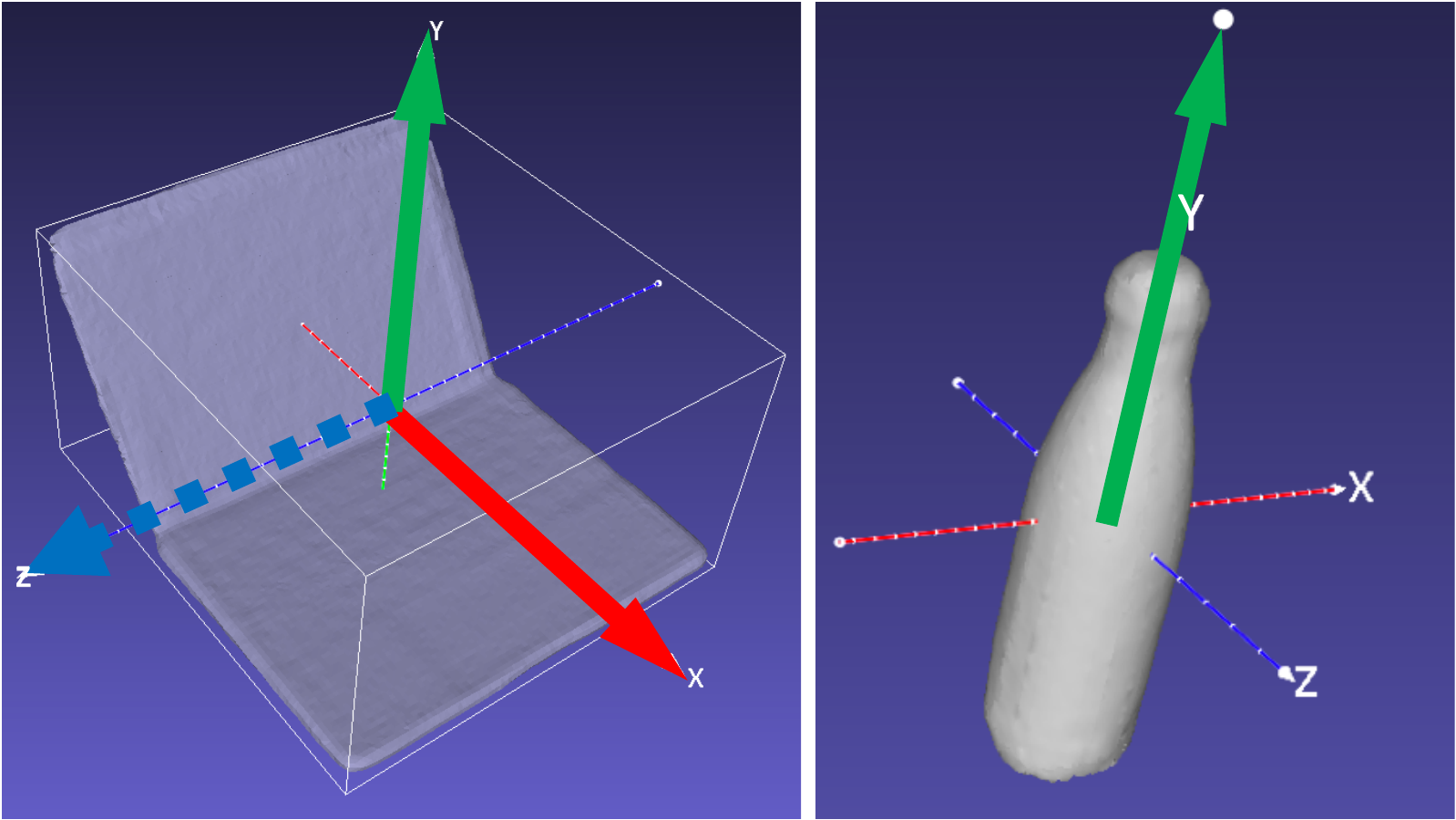}
\end{center}
   \caption{\textbf{Rotation represented by vectors.} Left: The object rotation can be represented by two perpendicular vectors (green vector and red vector); Right: For circular symmetry objects like the bottle, only the green(up) vector matters.}
\label{fig:axis_0}
\end{figure}

As shown in Fig. \ref{fig:axis_0}, we represent the rotation as two orthogonal vectors, called Flexible Vector-based Rotation (FVR).
In the following, we provide detailed analyse of the properties of the proposed FVR.
 
\subsubsection{Flexibility of FVR representation}
\label{sec:flex_fvr}
\begin{figure}[t!]
\begin{center}
\begin{tabular}{c}
{{\includegraphics[width=0.99\linewidth]{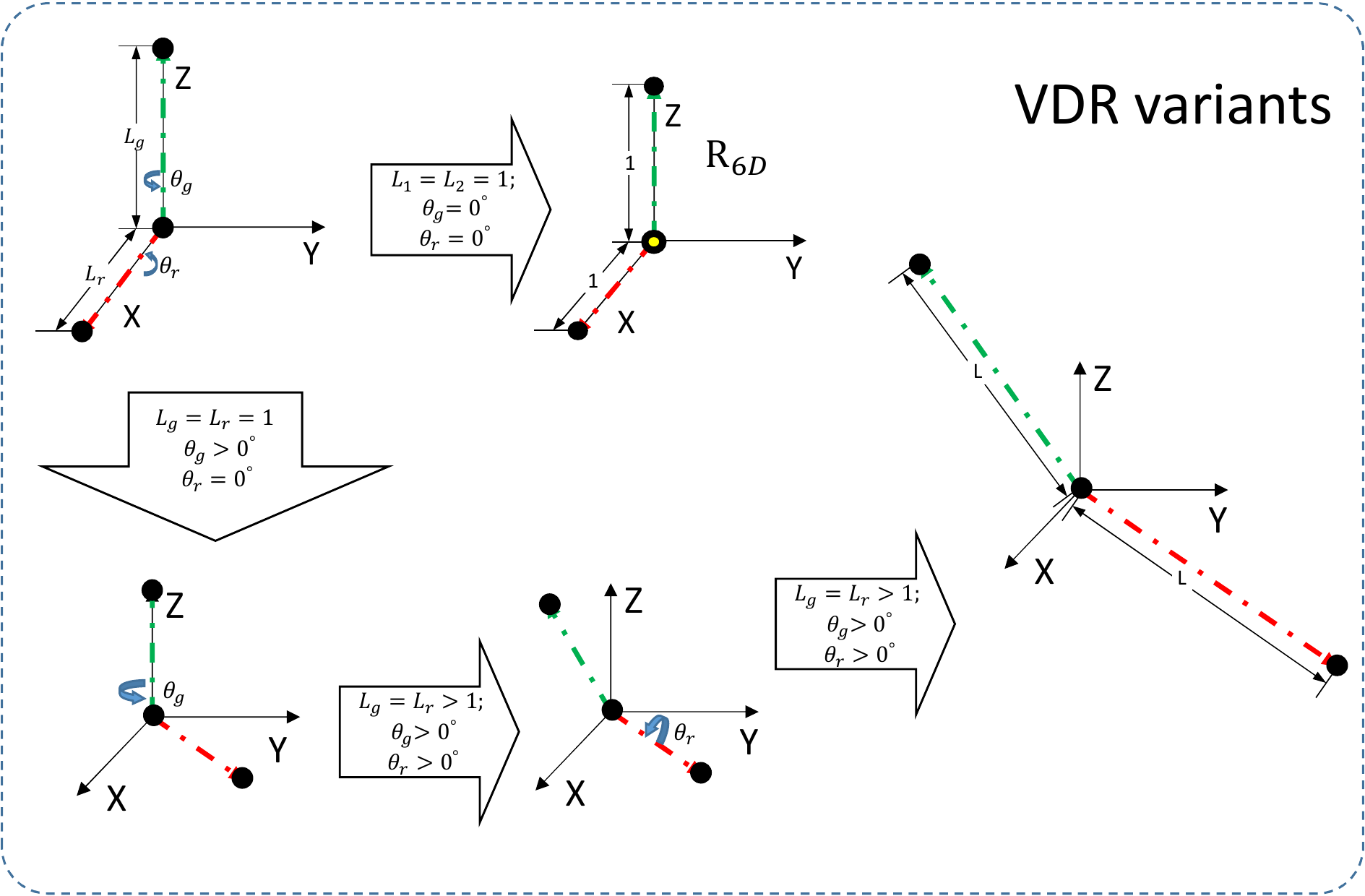}}}
\end{tabular}
\end{center}
\caption{\ronetwo{\textbf{FVR variants.} FVR consists of two orthogonal vectors which allows us to change the length and angles to generate different rotation presentations. R6D, however, is a special case of FVR by setting the angles to zeros and length to one.}}
\label{fig:vdr_6d}
\end{figure}

\ronetwo{FVR representation consists of two orthogonal vectors which can be parameterised as (shown in Fig. \ref{fig:vdr_6d}):
$V_{green}(\theta_r, L_g), V_{red}(\theta_g, L_r)$,
where $V_{green}$ and $V_{red}$ are orthogonal to each other. Without loss of generality, we assume that $\theta_r$ is angle that $V_{green}$ rotates around $V_{red}$, $\theta_g$ is angle that $V_red$ rotates around $V_{green}$. $L_g$ and $L_r$ are lengths of $V_{green}$ and $V_{red}$. In Fig. \ref{fig:vdr_6d}, we find that 
when we set $\theta_g =0, \theta_r =0, L_g=1, L_r=1$, then the endpoints of the transformed $V_{green}$ and $V_{red}$ are $R_{6D}$ representation (the two columns/rows of the rotation matrix), which means $R_{6D}$ is a variant of FVR.
While the flexibility of the proposed FVR representation allows us to change $\theta$s or $L$s to generate more suitable vector-based rotation variants for specific pose estimation task. Here we provide a simple solution to find the optimal parameters for the FVR representation. For simplicity, we set $L_g=L_r=L$, then we find the optimal value of $L$ and $\theta$s separately. 
We first search the optimal value of $L$ by fixing the value of $\theta_r$ and $\theta_g$, then we search the optimal value of $\theta_g$ by fixing the value of $\theta_r$ and $L$. Finally, we search the optimal value of $\theta_g$ by optimal $\theta_r$ and $L$. In Table \ref{tab:fsrots} and Table \ref{tab:g2lrots}, we report the object-wise and average rotation error with optimal parameters.
}

\subsubsection{Robust Vector-Fashion Estimation}
\begin{figure}[t!]
\begin{center}
\begin{tabular}{c}
{{\includegraphics[width=0.8\linewidth]{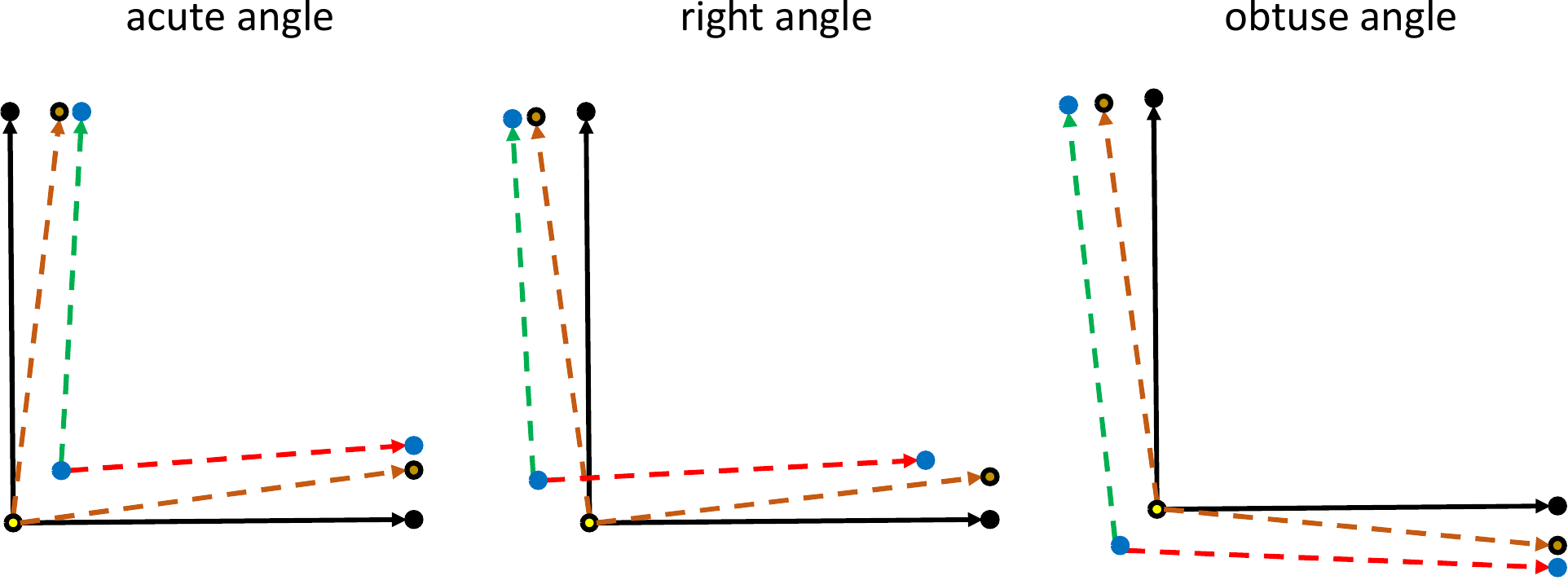}}}
\end{tabular}
\end{center}
\caption{\ronetwo{\textbf{Fluctuation cases}. The brown vectors denote the prediction of $R_{6D}$, and the blue and red vectors are FVR predictions. The black vectors are ground truth. From left to right: the predicted vectors fluctuation in acute, right, and obtuse cases. In the aspect of orientation, FVR prediction results are more close to ground truth, despite the bigger displacement error for starting points and endpoints.}}
\label{fig:fluc}
\end{figure}

\ronetwo{In $R_{6D}$ representation, the network is trained to predict the endpoints of the two transformed unit vectors. Since they use the fixed start points ($[0, 0, 0]$) to calculate the predicted rotation, thereby the estimation error of the endpoints will lead to the fluctuation of the vector orientation (shown in Fig. \ref{fig:fluc}), resulting in inaccurate calculation of the final rotation term.
 While our FVR representation allows the network to estimate both start point and endpoint of the vectors. As shown in Fig. \ref{fig:fluc}, since the start and end point of the vector are jointly adjusted, the proposed FVR is robust and less affected by the fluctuation of the endpoints when calculating the final rotation term. 
 From Table \ref{tab:fsrots}, we can see that our method outperforms $R_{6D}$ under both decoupled (9.48 vs 10.21) and non-decoupled (11.17 vs 11.40) setting with vector fashion training.}

\subsubsection{Decoupled Characteristic}
Since the two vectors are orthogonal, the rotation information related to one of them is independent to the other, which means that we can use one of them to recover partial rotation information of the object. For example, in Fig. \ref{fig:cate_show}, we use the green vector to recover the rotation for asymmetrical object. We can see that the green boxes and blue boxes are aligned well in the recovered vector orientation.

To decode the rotation information for different vectors, we utilise two decoders to extract the rotation information in a decoupled fashion. The two decoders decode the rotation information into the corresponding greed vector and red vector. Each decoder only needs to extract the orientation information along the corresponding vector which is easier than the estimation of the complete rotation. Although other rotation representations, such as Euler angle, quaternion, and axis-angle, consist of several parts and each part has its particular meaning, the estimation of different parts separately cannot achieve similar performance as the estimation of the whole representation (shown in Table \ref{tab:g2lrots}). One main reason is that the rotation information contained in sub-part of these representations is entangled with other sub-parts.

One evident advantage of this characteristic is that we can easily handle the symmetry objects like bottle, can, and bowl. Without loss of generality, we assume that the green vector is along the symmetry axis; then we can simply abandon the other vector while estimating the rotation to avoid rotation ambiguity.


\subsubsection{Point-Matching Variant}
Point-Matching loss is a widely used loss for pose estimation that calculates the loss based on ADD-(S) metric \cite{Hinterstoisser2013}. In \cite{Wang_2021_GDRN}, Wang et al. employed a new variant Point-Matching loss that accesses the loss only via rotation:
\begin{equation}
\mathcal{L}_{PM}=\operatorname{avg}_{\mathbf{x} \in \mathcal{M}}\|\hat{\mathbf{R}} \mathbf{x}-\overline{\mathbf{R}} \mathbf{x}\|_{2},
\end{equation}
where $\hat{\mathbf{R}}$ and $\mathbf{R}$ are the prediction and ground truth. $\mathcal{M}$ is the object 3D model.

\begin{figure}[t!]
\begin{center}
\begin{tabular}{cc}
{{\includegraphics[width=0.3\linewidth]{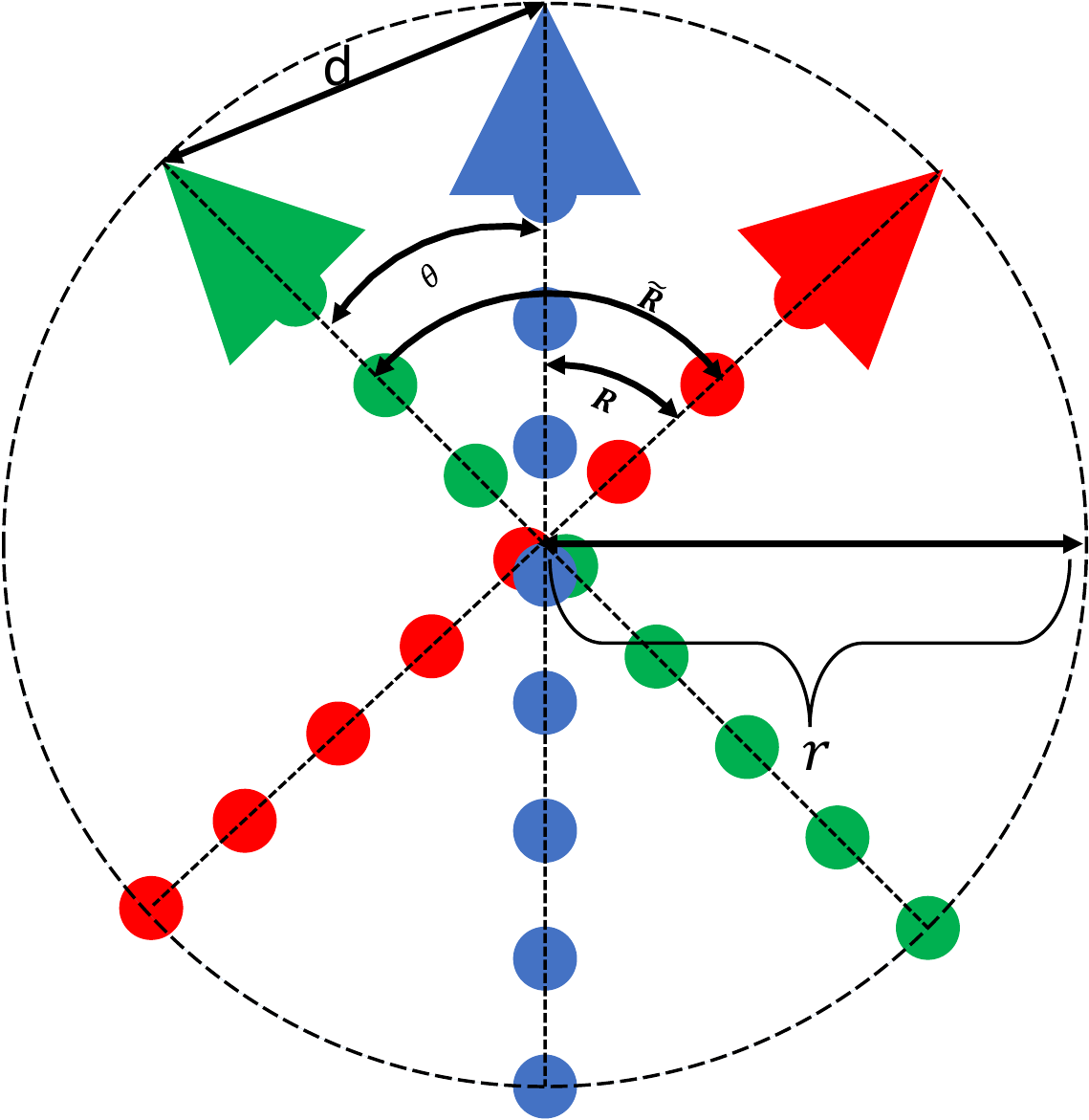}}}&
{{\includegraphics[width=0.3\linewidth]{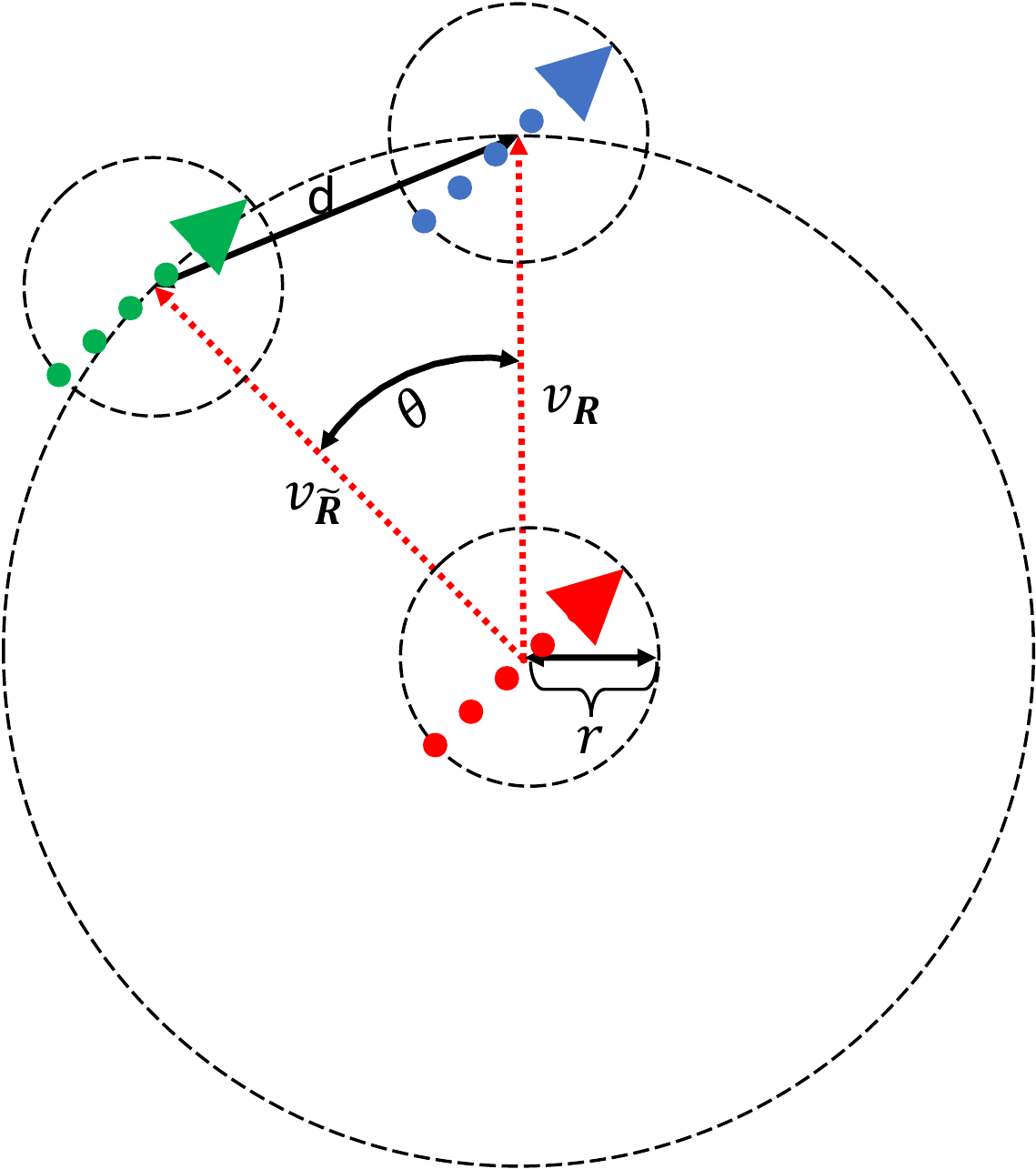}}}\\
(a) Point Matching \cite{Wang_2021_GDRN} & (b) Point Matching with FVR
\end{tabular}
\end{center}
\caption{\textbf{Point-Matching variant in 2D case.} Red point cloud arrow is the original point cloud. Blue is the one transformed by the ground truth value. Green is transformed by the estimated value. The left image describes the Point-Matching variant loss used in \cite{Wang_2021_GDRN} that only considers the rotation term. $r$ is the radius of the peripheral circle. $d$ is the distance between the point in the blue point cloud transformed by ground truth rotation $\textbf{R}$ and the point in the green point cloud transformed by estimated rotation $\widetilde{\textbf{R}}$. $\theta$ is the rotation residual between estimated rotation and the ground truth. The right image describes the Point-Matching variant based on our proposed FVR representation where the point clouds are transformed by the orientation of the vectors.}
\label{fig:drpm}
\end{figure}

However, there exists one issue here. As shown in Fig. \ref{fig:drpm}.a, when the rotation angle $\theta$ changes, the distance $d$ changes as well, their relation can be formalized as:
\begin{equation}
	d = 2r\sin(\frac{\theta}{2})
\label{eq:pm6d}
\end{equation}

From the Eq. \ref{eq:pm6d}, we can see that the distance is determined by rotation angle $\theta$ and the radius $r$ of the peripheral circle. When the point is closer to the center of rotation, the distance is shorter, which means the points away from the center are more important to the loss value. However, the conventional Point-Matching loss does not consider this imbalance problem.
%
%

To address this issue, we propose to use our FVR representation for Point-Matching loss. Since our new representation is vector-based, we cannot directly calculate the loss by multiplying the FVR with the object 3D model points. Therefore, we use the add operation to replace the multiplication in Eq. \ref{eq:pm6d}. As shown in Fig. \ref{fig:drpm}.b, we move the 3D object points to the direction pointed by the rotation vectors. Then the new Point-Matching loss can be defined as:
\begin{equation}
\begin{split}
		\mathcal{L}_{\mathbf{R}} = \operatorname{avg}_{\mathbf{x} \in \mathcal{M}}\|\mathbf{x}+{\mathbf{v}}_g-(\mathbf{x}+\hat{\mathbf{v}}_g)\|_2\\
		+\operatorname{avg}_{\mathbf{x} \in \mathcal{M}}\|\mathbf{x}+{\mathbf{v}}_r-(\mathbf{x}+\hat{\mathbf{v}}_r)\|_2
\end{split}
\label{eq:pmr}
\end{equation}
where $\hat{\textbf{v}}_g$ and $\hat{\textbf{v}}_r$ are the predicted green and red vectors. $\textbf{v}_g$ and $\textbf{v}_r$ are the ground truth.

It is obvious that the new Point-Matching loss $\mathcal{L}_{\mathbf{R}}$ is equal to mean squared error (MSE) loss between the prediction and ground truth FVR:
\begin{equation}
\begin{split}
		\mathcal{L}_{\mathbf{R}}
		=\operatorname{avg}_{\mathbf{x} \in \mathcal{M}}\|{\mathbf{v}}_g-\hat{\mathbf{v}}_g\|_2\\
		+\operatorname{avg}_{\mathbf{x} \in \mathcal{M}}\|{\mathbf{v}}_r-\hat{\mathbf{v}}_r\|_2\\
		=\|\mathbf{v}_g-\hat{\mathbf{v}}_g\|_2+\|{\mathbf{v}}_r-\hat{\mathbf{v}}_r\|_2
\end{split}
\label{eq:mser}
\end{equation}

\noindent That means with MSE loss, the proposed FVR representation can naturally avoid the imbalance issue existing in Eq. \ref{eq:pm6d}.



\subsection{Residual Prediction Network}
\label{sec:res_net}
As both translation and object size are related to points coordinates, inspired by \cite{Qi_2018_CVPR,Chen_2020_CVPR}, we train a tiny PointNet \cite{Qi_2017_CVPR} that takes segmented point cloud as input. More concretely, the PointNet performs two tasks: 1) estimating the residual between the translation ground truth and the mean value of the segmented point cloud; 2) estimating the residual between object size and the mean category size. 

For size residual, we pre-calculate the mean size $[\overline{x}, \overline{y}, \overline{z}]^T$ of each category.
Then for object $o$ in that category the ground truth $[\delta_x^o,
\delta_y^o,
\delta_z^o]^T$ of the size residual estimation is calculated as
\begin{equation}
[\delta_x^o,
\delta_y^o,
\delta_z^o]^T = [x_o,
y_o, 
z_o]^T - [
\overline{x},
\overline{y},
\overline{z}]^T.
\end{equation}

We use MSE loss to predict both the translation and size residual. The total loss function $\mathcal{L}_{res}$ is defined as: $\mathcal{L}_{res} = \mathcal{L}_{tra}+\mathcal{L}_{size}$,
where $\mathcal{L}_{tra}$ and $\mathcal{L}_{size}$ are sub-loss for translation residual and size residual, respectively. 
\rone{In Table \ref{tab:abla}, we show the effect of residual prediction network for pose estimation task.}

\subsection{Data Augmentation: 3D Deformation Mechanism}
One major issue in category-level 6D pose estimation is the intra-class shape variation. The existing methods employed two large synthetic datasets, i.e., CAMERA275(NOCS-Synthetic) \cite{wang2019nocs} and 3D model dataset \cite{chang2015shapenet} to learn this variation. 
However, this strategy not only needs extra hardware resources but also increases the (pre-)training time.

\begin{figure}[t!]
\begin{center}
\includegraphics[width=0.99\linewidth]{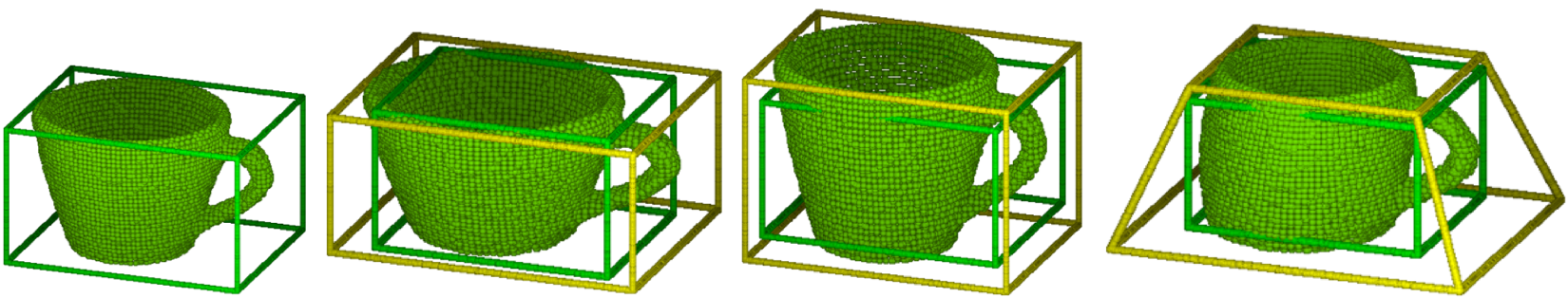}

\end{center}
\hspace{-10pt}
\vspace{-10pt}
\caption{\textbf{3D deformed examples}. The new training examples can be generated by enlarging, shrinking, or changing the area of some surfaces of the box-cages. 
The left one is the original point could with the original 3D box-cage, i.e. 3D bounding box. The right three ones are the deformed point clouds with deformed box-cages (shown in yellow color). The green boxes are the original 3D bounding boxes before deformation.}
\label{fig:3d_defor}
\end{figure}

To alleviate the shape variation issue, based on the fact that the shapes of most objects in the same category are similar \cite{vlach2016we} (shown in Fig. \ref{fig:color_var}), we propose an online 3D deformation mechanism for training data augmentation. 
We pre-define a box-cage for each rigid object (shown in Fig. \ref{fig:3d_defor}). Each point is assigned to its nearest surface of the cage; when we deform the surface, the corresponding points move as well.

Though box-cage can be designed more refined, in experiments, we find that with a simple box cage, i.e. 3D bounding box of the object, the generalization ability of FS-Net is considerably improved (Table \ref{tab:abla}). 
Different to \cite{yifan2020neural}, we do not need an extra training process to obtain the box-cage of the object, and we do not need target shape to learn the deformation operation either. Our mechanism can be applied during training on the fly which can save training time and storage space.

To make the deformation operation easier, we first transfer the points to the canonical coordinate system and then perform 3D deformation. Finally, we transform them back to the global scene:
\begin{equation}
    \{\mathcal{P}_1,\mathcal{P}_2,\cdots ,\mathcal{P}_n\}=R(\mathbb{F}_{3D}(R^T(\mathcal{P}-T)))+T,
\end{equation}
where $\mathcal{P}$ is the points generated after the 2D detection step. $R$, $T$ are the pose ground truth. $\{\mathcal{P}_1,\mathcal{P}_2,\cdots ,\mathcal{P}_n\}$ are the new generated training examples. $\mathbb{F}_{3D}$ is 3D deformation which includes cage enlarging, shrinking, changing the area of some surfaces. In Section \ref{sec:gen_per}, we show how the propose 3D deformation mechanism can increase the generalization ability of FS-Net.

\section{Experiments}


\subsection{Datasets}
\noindent \textbf{NOCS-REAL} \cite{wang2019nocs} is the first real-world dataset for category-level 6D object pose estimation. The training set has 4300 real images of 7 scenes with 6 categories. For each category, there are 3 unique instances. In the testing set, there are 2750 real images spread in 6 scenes of the same 6 categories as the training set. In each test scene, there are about 5 objects which makes the dataset clutter and challenging.

\noindent \textbf{LINEMOD} \cite{Hinterstoisser2013} is a widely used instance-level 6D object pose estimation dataset which consists of 13 different objects with significant shape variation.

However, the original 6D pose estimation datasets do not provide point-wise label. To train the proposed network in a supervised way, we use the method proposed in our previous work \cite{Chen_2020_WACV} to generate the point-wise label.

\subsection{Training Details}
We use Pytorch to implement our pipeline. All experiments are deployed on a PC with i7-4930K 3.4GHz CPU and GTX 1080Ti GPU.

First, to locate the object in RGB images, we fine-tune the YOLOv3 pre-trained on COCO dataset \cite{lin2014microsoft} with the training dataset.
Then we jointly train the 3DGC autoencoder, rotation estimation and residual estimation network. The total loss function is defined as:
\begin{equation}
\mathcal{L}_{Shape} =  \lambda_{seg} \mathcal{L}_{seg}+\lambda_{rec} \mathcal{L}_{rec} +\lambda_{\mathbf{R}} \mathcal{L}_{\mathbf{R}} +\lambda_{res} \mathcal{L}_{res},
\end{equation}
where $\lambda$s are the balance parameters which are set to keep different loss values at the same magnitude. We use cross entropy for 3D segmentation loss function $\mathcal{L}_{seg}$.

We adopt Adam \cite{kingma2014adam} to optimize the FS-Net. The initial learning rate is 0.001, and it is halved every 10 epochs. The maximum epoch is 50. During inference, given a 640$ \times$ 480 RGB-D image, our FS-Net runs at 20 FPS on Intel i7-4930K CPU and 1080Ti GPU. Specifically, the 2D detection takes about 10ms to proceed. The pose and size estimation takes about 40ms.
\subsubsection{Gap Between Different Architectures}
In the test phase, our method first uses the color image to generate a bounding box to locate the corresponding position of the depth image, and then convert the cropped depth image into point cloud for subsequent processing. 
However, in the training stage, the point cloud is generated from the ground truth bounding box. When we only use the points under these boxes, the generalization performance of the trained model will suffer(see Table \ref{tab:catepose}). 
The reason is that there exist some gap between the ground truth and prediction, while the previous methods \cite{wang2019nocs, tian2020shapeprior} did not consider how to mitigate the differences which may lead to overfitting issue of the trained model.
One can generate the data with various bounding boxes or masks before training or during training, while these strategies will significantly increase the storage burden or training time. Here we provide an efficient way to solve this issue.

\subsubsection{Fast Point-Wise Relabelling}
\begin{figure}[t!]
\begin{center}
\begin{tabular}{cccc}
{{\includegraphics[width=2.5cm]{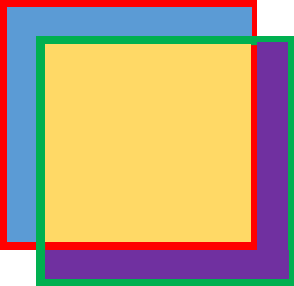}}}&
{{\includegraphics[width=2.5cm]{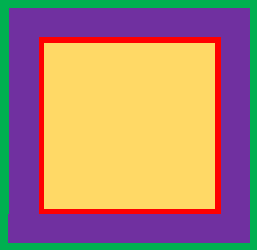}}}&
{{\includegraphics[width=2.5cm]{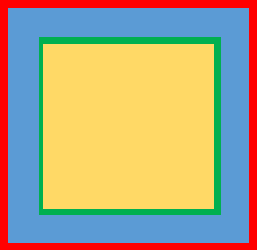}}}\\
\end{tabular}
\end{center}
\vspace{-5pt}
\caption{\textbf{Boxes relations}. Green 2D bounding box denotes ground truth. Red bounding box represents augmented box. The yellow region is the intersection of the two boxes; the purple region is the region in ground truth box but outside the intersection region; the blue region is the region that belongs to the augmented box apart from the intersection region.}
\label{fig:rela_2d}
\end{figure}

From Fig. \ref{fig:rela_2d}, we can see that there are three valid regions for the ground truth and augmented 2D boxes: intersection region (yellow), newly added region (blue), and outside augmented region (purple). Our task here is to access the point-wise label in the blue and yellow region, i.e., the augmented 2D box.


Our solution is labelling the yellow region and blue region separately. First, we find the intersection region of the augmented and ground truth box (yellow region shown in Fig. \ref{fig:rela_2d}) and set the points in that region as object points. Then we set the depth value in the yellow region as zero, thereby, for the transformed point cloud the coordinate of $z$ is zero. By removing the points with $z$ zero, we can access the points that are located in the blue region. Finally, we label these points in the blue region as background points.

\subsection{Evaluation Metrics}
For category-level pose estimation, we adopt the same metrics  $IoU_{X}$  and $n^{\circ}$ $m$ $\textbf{cm}$ used in \cite{wang2019nocs, chen2020cass,tian2020shapeprior}:
\begin{itemize}
\item$IoU_{X}$ is Intersection-over-Union (IoU) accuracy for 3D object detection under different overlap thresholds. The overlap ratio larger than $X$ is accepted.

\item $n^{\circ}m\textbf{cm}$ represents pose estimation error of rotation and translation. The rotation error less than $n^{\circ}$ and the translation error less than $m\textbf{cm}$ is accepted.
\end{itemize}

For instance-level pose estimation, we compare the performance of FS-Net with other state-of-the-art instance-level methods using the ADD-(S) metric \cite{Hinterstoisser2013}.

\subsection{Ablation Studies}
\begin{table}[t]
\caption{\textbf{Ablation studies on NOCS-REAL dataset}. We use two different metrics to measure performance. `3DGC' means the 3D graph convolution. `OPR' means observed points reconstruction. `FVR' represents the decoupled rotation mechanism. `DEF' denotes the online 3D deformation. Please note, for the sake of the ablation study, we provide the ground truth 2D bounding box for different methods.}
\label{tab:abla}
\vspace{2mm}
\centering
\resizebox{0.99\linewidth}{12mm}
{
\begin{tabular}{|l|cccccccc|}
\hline
Method & 3DGC& DEF &  OPR  & FVR& $IoU_{50}$ & \rtwo{$IoU_{90}$} & 10$^{\circ}10cm$& \rtwo{2$^{\circ}$2 \textbf{cm}}\\
\hline
G2L \cite{Chen_2020_CVPR} & $\times$ &$\checkmark$& $\times$ & $\times$& 94.65\% & 19.87\% &31.0\% & 2.87\%\\
G2L+FVR & $\times$ &$\checkmark$& $\times$ & $\checkmark$& 96.21\% &27.91\% &47.81\% &8.11\%\\
Med1 &$\checkmark$ &$\checkmark$ &  $\times$ &  $\times$& 97.98\% &22.11\% &46.4\% &4.10\%\\
Med2 & $\checkmark$&$\checkmark$  & $\checkmark$ & $\times$ &95.61\% &14.93\% &46.8\%&4.26\%\\
Med3 & $\checkmark$ &$\checkmark$ &  $\times$ & $\checkmark$& 97.34\%&32.18\% &61.1\%&14.68\%\\
Med4 & $\checkmark$ &$\times$ & $\checkmark$ & $\checkmark$&97.30\% &25.08\% & 58.2\%&12.77\%\\
Med5(Ours) & $\checkmark$ &$\checkmark$ & $\checkmark$ & $\checkmark$&98.04\% &32.51\% &65.9\%&16.05\% \\
Med6 & $\checkmark$ &$\checkmark$ & whole model & $\checkmark$&94.44\% &27.31\% &58.0\%&  15.07\% \\
\rone{w/o residual} & $\checkmark$ &$\checkmark$ & $\checkmark$ & $\checkmark$&48.72\% &0.0\% &21.2\%&  3.13\% \\
\hline
\end{tabular}
}
\end{table}




We use the G2L-Net \cite{Chen_2020_CVPR} as the baseline method which extracted the latent feature for rotation estimation via point-wise orientated vector regression, and the ground truth of rotation is the eight corners of 3D bounding box with the corresponding rotation. The loss function for rotation estimation is the MSE loss between predicted 3D coordinates and ground truth.
Compared to baseline, our proposed work has three novelties: a) view-based 3DGC autoencoder for observed point cloud reconstruction; b) FVR representation; c) 3D augmentation mechanism.

In Table \ref{tab:abla}, we report the experimental results of three novelties on the NOCS-REAL dataset. Comparing Med3 and Med5, we find that reconstruction of the observed point cloud can learn better pose feature. The performance of Med2(Med1, G2L) and Med5(Med3, G2L+FVR) shows that the proposed FVR representation can effectively extract the rotation information. The results of Med4 and Med5 demonstrate the effectiveness of the 3D deformation mechanism, which increases the pose accuracy by $7.7\%$ in terms of 10$^{\circ}10cm$ metric.
We also compare the different reconstruction choices: the reconstruction of observed points and the complete object model with corresponding rotation. Compared Med5 and Med6, we can see that the observed points reconstruction can learn better rotation feature than whole object model reconstruction. Overall, Table \ref{tab:abla} shows that the proposed novelties can improve the accuracy significantly. 

\subsection{Generalization Performance}
\label{sec:gen_per}
\begin{figure}[htp!]
\begin{center}
\includegraphics[width=0.7\linewidth]{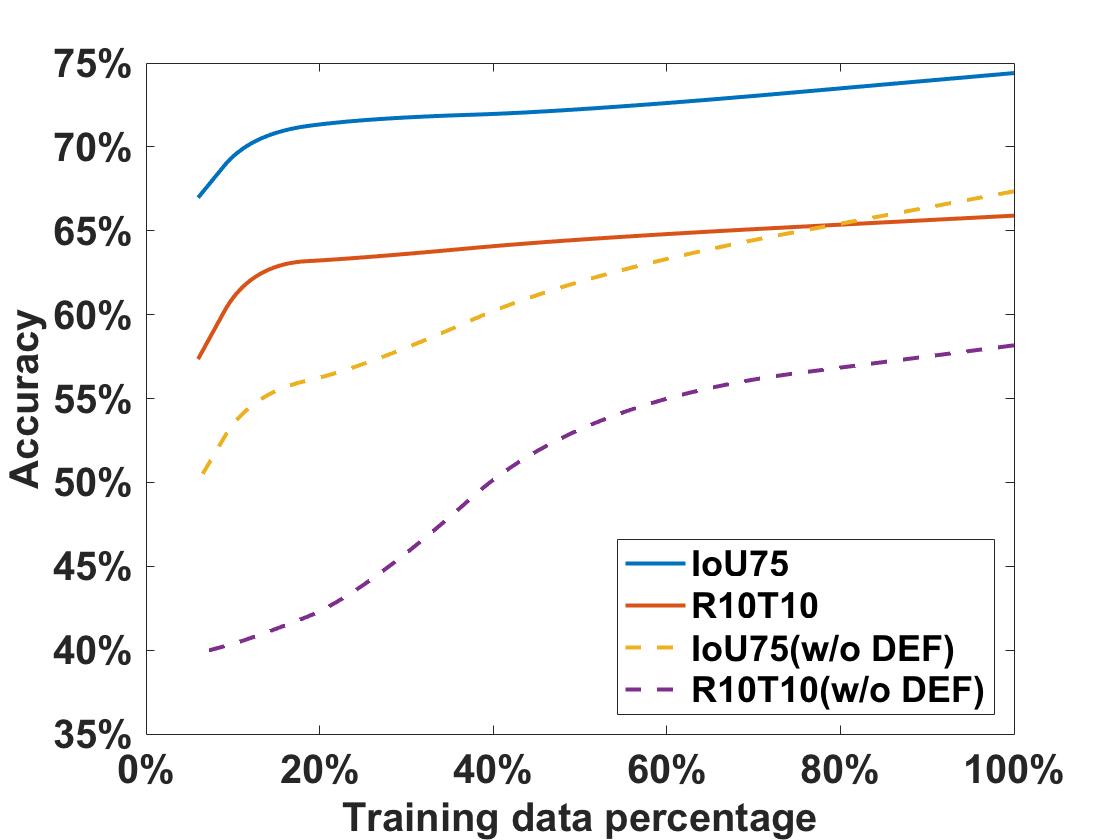}

\end{center}
  \caption{\textbf{Generalization performance.} With the given 2D bounding box and a randomly chosen 3D sphere center, we show how the training set size affects the pose estimation performance. `w/o DEF' means no 3D deformation mechanism is adopted during training.}
\label{fig:per}
\end{figure}

To test the generalization performance of FS-Net, we randomly choose different percentages of the training set to train FS-Net and test it on the whole testing set. Fig. \ref{fig:per} shows that: 1) FS-Net is robust to the size of the training dataset. Even with $20\%$ of the training dataset, the FS-Net can still achieve good performance; 2) the 3D deformation mechanism significantly improves the robustness and performance of FS-Net.


\subsection{Comparison with State-of-the-Arts}

\subsubsection{Category-Level Pose Estimation}
\begin{figure*}[htp!]
\begin{center}
\includegraphics[width=0.9\linewidth]{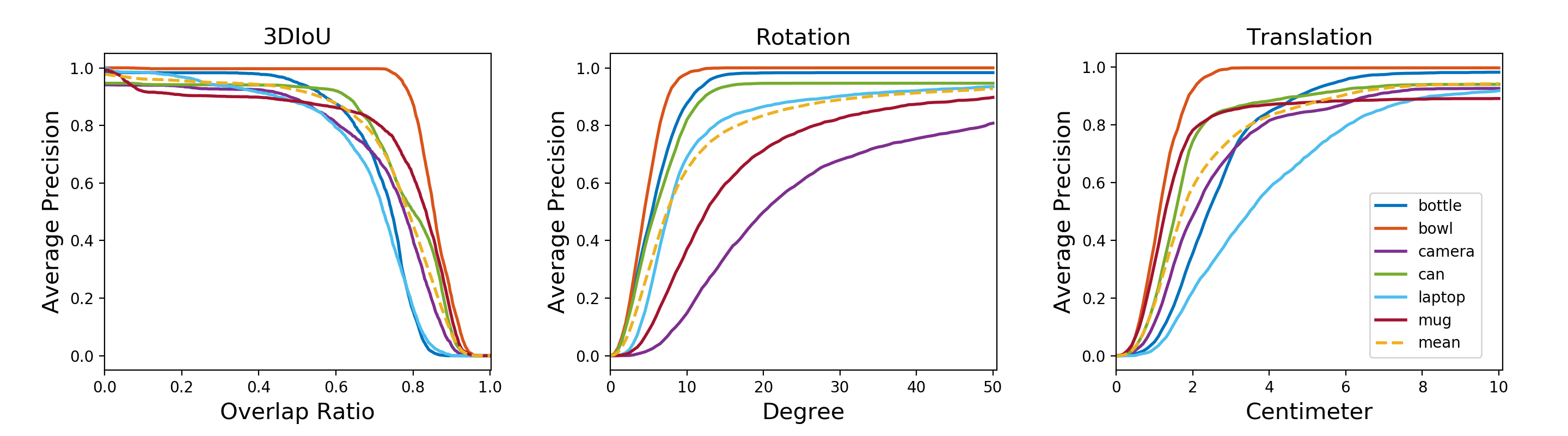}
\caption{\textbf{Result on NOCS-REAL}. The average precision of different thresholds tested on NOCS-REAL with 3D IoU, rotation, and translation error.}
\label{fig:iourt}
\end{center}
\vspace{-10pt}
\hspace{-10pt}
\end{figure*}

\begin{figure*}[ht!]
\begin{center}
\resizebox{0.99\textwidth}{35mm}{
\begin{tabular}{cccccc}
{{\includegraphics[width=0.15\linewidth]{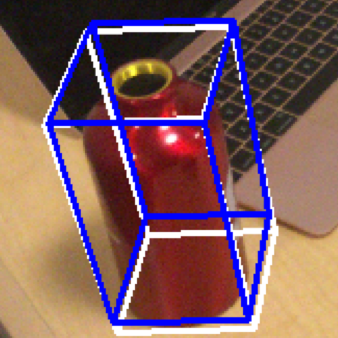}}}&
{{\includegraphics[width=0.15\linewidth]{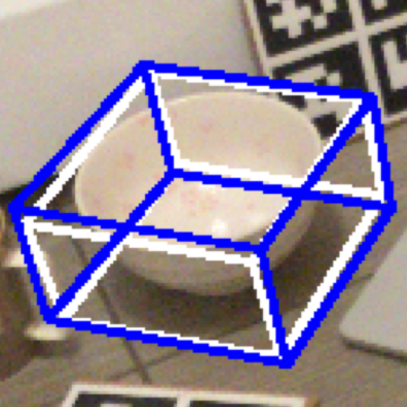}}}&
{{\includegraphics[width=0.15\linewidth]{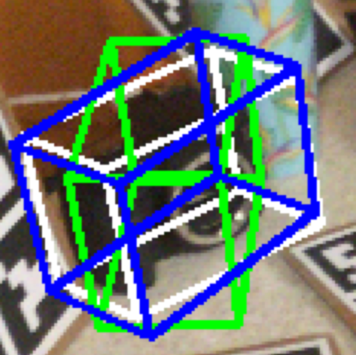}}}&
{{\includegraphics[width=0.15\linewidth]{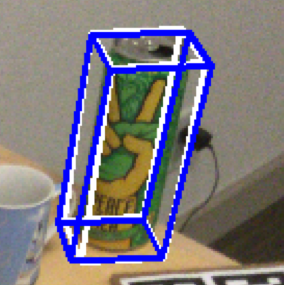}}}&
{{\includegraphics[width=0.15\linewidth]{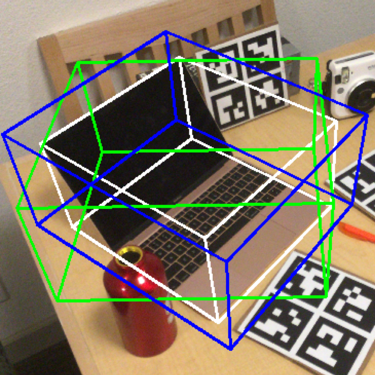}}}&
{{\includegraphics[width=0.15\linewidth]{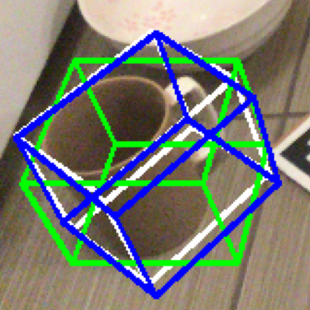}}}\\
{{\includegraphics[width=0.15\linewidth]{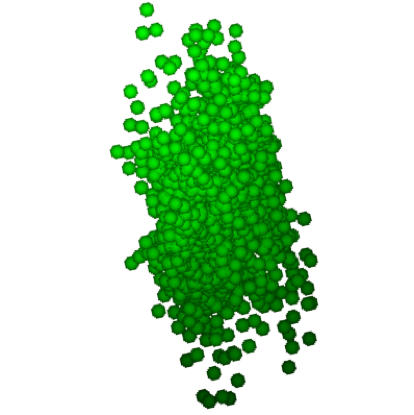}}}&
{{\includegraphics[width=0.15\linewidth]{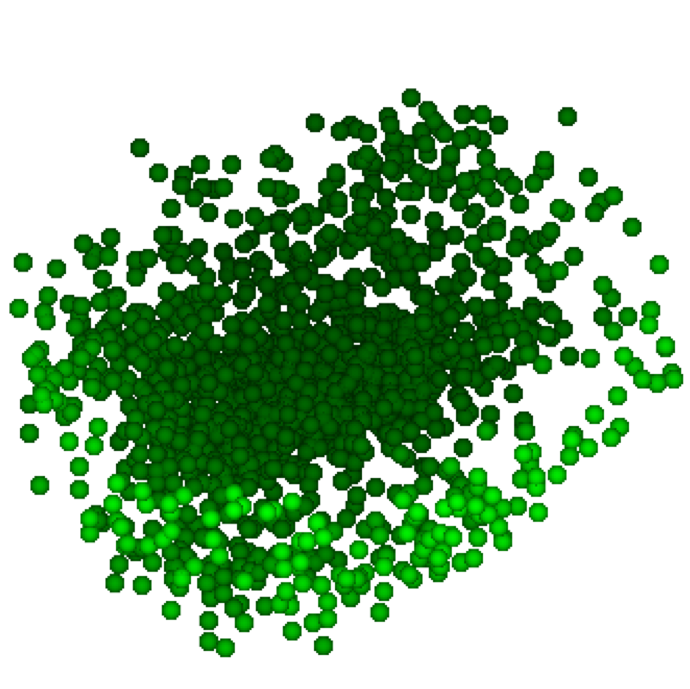}}}&
{{\includegraphics[width=0.15\linewidth]{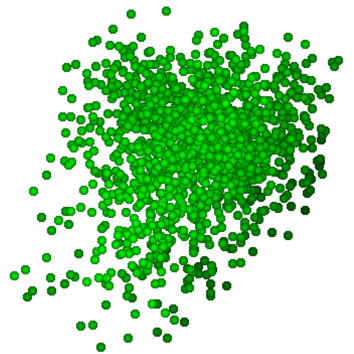}}}&
{{\includegraphics[width=0.15\linewidth]{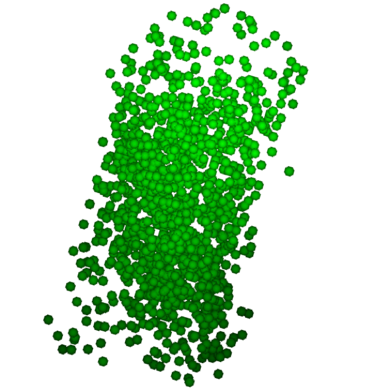}}}&
{{\includegraphics[width=0.15\linewidth]{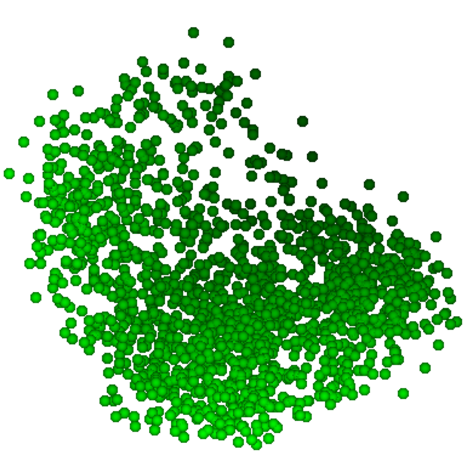}}}&
{{\includegraphics[width=0.15\linewidth]{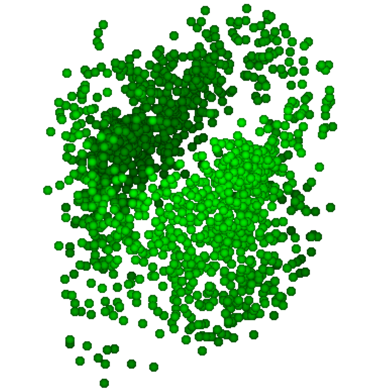}}}\\
{{\includegraphics[width=0.15\linewidth]{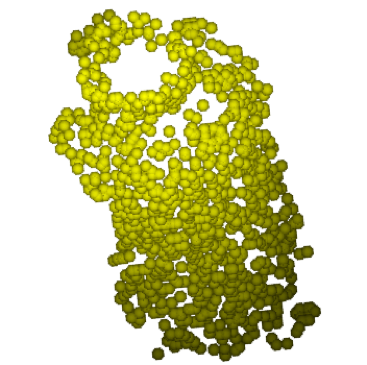}}}&
{{\includegraphics[width=0.15\linewidth]{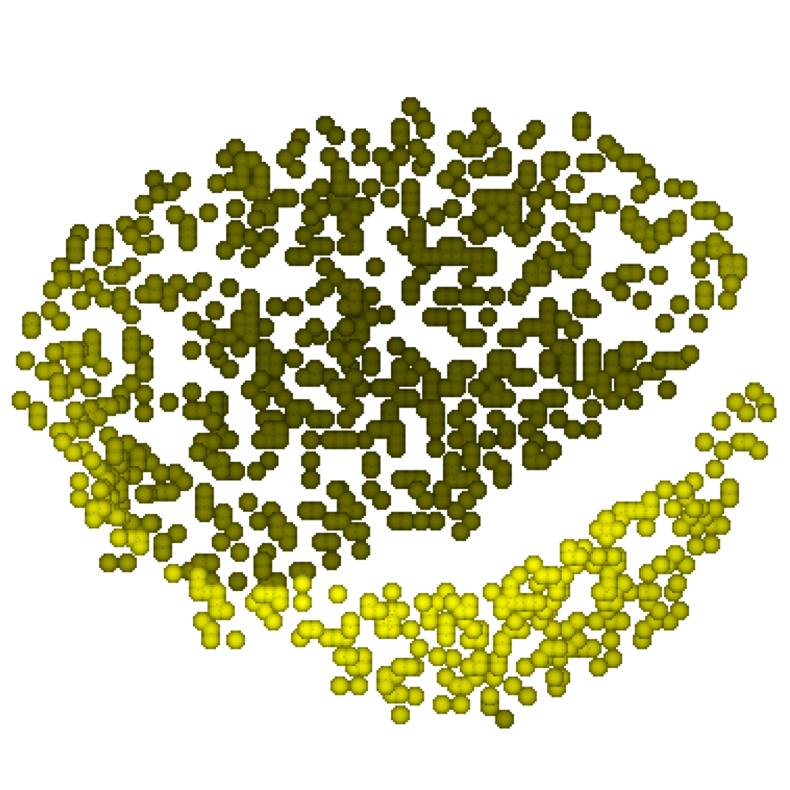}}}&
{{\includegraphics[width=0.15\linewidth]{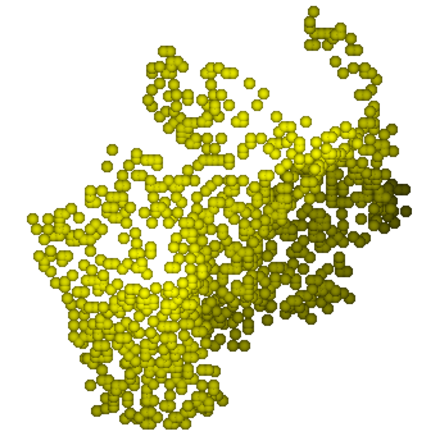}}}&
{{\includegraphics[width=0.15\linewidth]{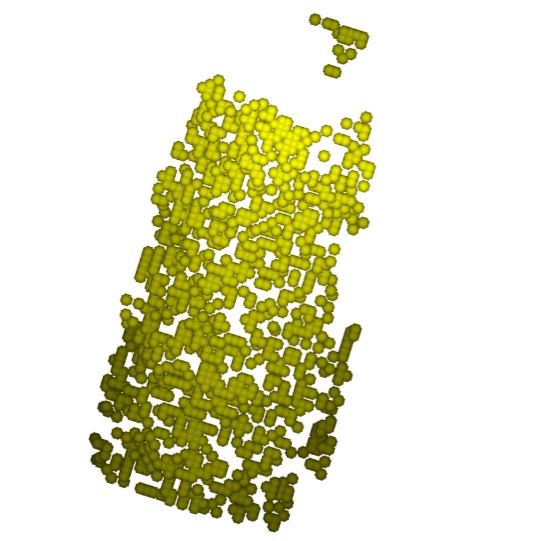}}}&
{{\includegraphics[width=0.15\linewidth]{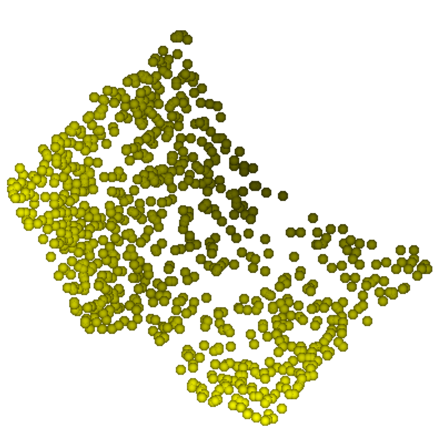}}}&
{{\includegraphics[width=0.15\linewidth]{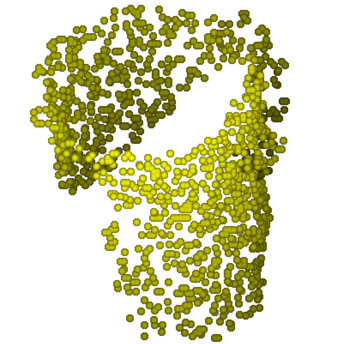}}}\\
\end{tabular}}
\end{center}
\vspace{-10pt}
\hspace{-10pt}
\caption{\textbf{Qualitative results on NOCS-REAL dataset}. The first row is the pose and size estimation results. White boxes denote ground truth. Blue boxes are the poses recovered from two rotation vectors. The green boxes are the poses recovered from the green vector \rone{(since `bottle', `bowl' and `can' only need green vector to recover pose, we only show blue boxes of asymmetrical objects)}. Our results match ground truth well in both pose and size. The second row is the reconstructed observed points under corresponding poses. The third row is the ground truth of the observed points transformed from the observed depth map.}
\label{fig:cate_show}
\end{figure*}

We compare FS-Net with NOCS \cite{wang2019nocs}, CASS \cite{chen2020cass}, Shape-Prior \cite{tian2020shapeprior}, and 6-PACK \cite{wang20206dpack} on NOCS-REAL dataset in Table \ref{tab:catepose}. We can see that our proposed method outperforms the other state-of-the-art methods on both accuracy and speed. Specifically, with 3D detection metric $IOU_{50}$, our FS-Net outperforms the previous best method, NOCS, by $11.7\%$ and the running speed is 4 times faster. In terms of 6D pose metric 5$^{\circ}5cm$ and 10$^{\circ}10cm$, FS-Net outperforms the CASS by the margins of $4.7\%$ and $6.3\%$, respectively. FS-Net even outperforms 6-PACK under 3D detection metric $IOU_{50}$, which is a 6D tracker and needs an initial 6D pose and object size to start. See Fig. \ref{fig:iourt} for more quantitative details. \rone{In addition, we train our FS-Net with both synthetic and real data. From Table \ref{tab:catepose}, we can see that the performance further improved, and is comparable with the state-of-the-art methods SGPA \cite{chen2021sgpa} and DualPoseNet \cite{lin2021dualposenet}.}
The qualitative results are shown in Fig. \ref{fig:cate_show}. From Fig. \ref{fig:cate_show} we can see that although the reconstructed points are not perfectly in line with the target points, the basic orientation information is kept.
 Please note, we only use real-world data (NOCS-REAL) to train our pose estimation part. Other methods use both synthetic dataset (CAMERA) \cite{wang2019nocs} and real-world data for training. The number of training examples in the CAMERA is 275K, which is more than 60 times than that of NOCS-REAL (4.3K). It shows that FS-Net can efficiently extract the category-level pose feature with fewer data.


\begin{table}[htb]

\caption{\textbf{Instance-level comparison on LINEMOD dataset.} Our method achieves a comparable performance with the state-of-the-art in both speed and accuracy.}
\label{tab:linmod}
\vspace{1mm}
\centering

\resizebox{0.4\textwidth}{!}{
\begin{tabular}{lccc}
\hline
Method & Input & ADD-(S) & Speed(FPS)\\\hline

PVNet \cite{peng2018pvnet}& RGB &  86.3\% & 25 \\
CDPN \cite{li2019cdpn}& RGB &  89.9\% & 33 \\
DPOD \cite{zakharov2019dpod}& RGB &  95.2\% & 33 \\
G2L-Net \cite{Chen_2020_CVPR} & RGBD &  98.7\% & 23 \\
Densefusion\cite{wang2019densefusion} & RGBD &  94.3\% & 16  \\
PVN3D \cite{he2020pvn3d} &RGBD&  99.4\% &5 \\
Ours & RGBD&  {97.6\%} & 20 \\
\hline
\end{tabular}}
\end{table}

%
%
%
%

\begin{table}[htb!]
\centering
\caption{\textbf{Category-level performance on NOCS-REAL dataset.} We summarize the pose estimation results reported in the origin papers on the NOCS-REAL dataset. `-' means no results are reported under this metric. The values in the bracket are the results for synthetic NOCS dataset. $\ast$ means that the work is published after the submission of this manuscript.}
\label{tab:catepose}
\resizebox{0.49\textwidth}{12mm}
{
\begin{tabular}{lccccccc}
\hline
Method &$IoU_{25}$ & $IoU_{50}$ & $IoU_{75}$ & 5$^{\circ}5cm$& 10$^{\circ}5cm$& 10$^{\circ}10cm$& Speed(FPS)\\
\hline
NOCS \cite{wang2019nocs} & 84.9\%&  80.5\% & 30.1\% &  9.5 \% & 26.7\%& 26.7\%&  5 \\
CASS \cite{chen2020cass} & 84.2\%&  77.7\% & - &  23.5 \% & 58.0\%& 58.3\%&   - \\
Shape-Prior \cite{tian2020shapeprior} &83.4\%&  77.3\% & 53.2\% &  21.4\% & 54.1\%& -&  4\\
6-PACK \cite{wang20206dpack} & 94.2\%&  - & - &  {33.3 \%} & -& -&  10\\

\rone{$\ast$ CAPTRA \cite{Weng_2021_ICCV}} & -&  - & - &  { 63.6\%} & & -&  12\\

\rone{$\ast$ SGPA \cite{chen2021sgpa}} & -&  80.1\% & 61.9\% &  {39.6 \%} & 70.7\%& -&  -\\
\rone{$\ast$ DualPoseNet \cite{lin2021dualposenet}} & -&  79.8\% & 62.2\% &  {35.9 \%} & 66.8\%& -&  -\\\hline

Ours & {95.1\%}&  {92.2\%} & {63.5\%}&  {28.2 \%} & {60.8\%}& {64.6\%}&  {20}\\

Ours(with syn) & {95.2\%}&  {92.4\%} & {65.8\%}&  {35.4 \%} & {62.5\%}& {68.6\%}&  {20}\\
\hline
\end{tabular}
}
\end{table}

\subsubsection{Instance-Level Pose Estimation}
We compare the instance-level pose estimation results of FS-Net on the LINEMOD dataset with other state-of-the-arts instance-level methods. \rone{ Following the experimental protocol in G2L-Net \cite{Chen_2020_CVPR}, we set the initial learning rate as 0.001 and halve it every 50 epochs. The maximum epoch is 200.}

From Table \ref{tab:linmod}, we can see that FS-Net achieves comparable results on both accuracy and speed. It shows that our method can effectively extract both category-level and instance-level pose features. \rone{Please note that we do not apply 3D augmentation in instance-level pose estimation, since the same objects are shared in both training set and testing set}.

\begin{figure*}[htp!]
\centering
\begin{tabular}{ccc}
{\includegraphics[width=0.3\linewidth]{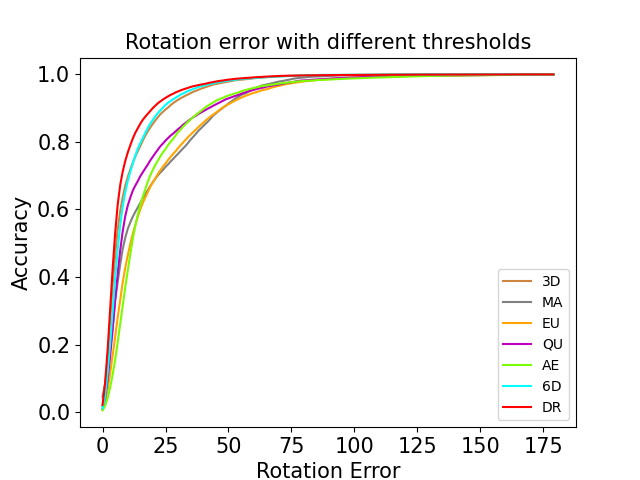}}&
{\includegraphics[width=0.3\linewidth]{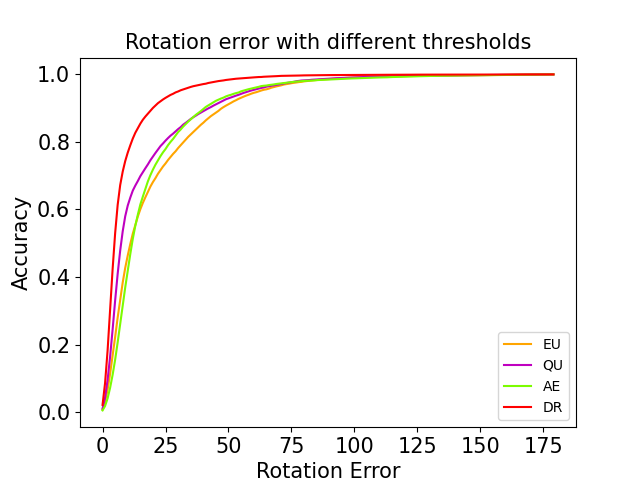}}&
{\includegraphics[width=0.3\linewidth]{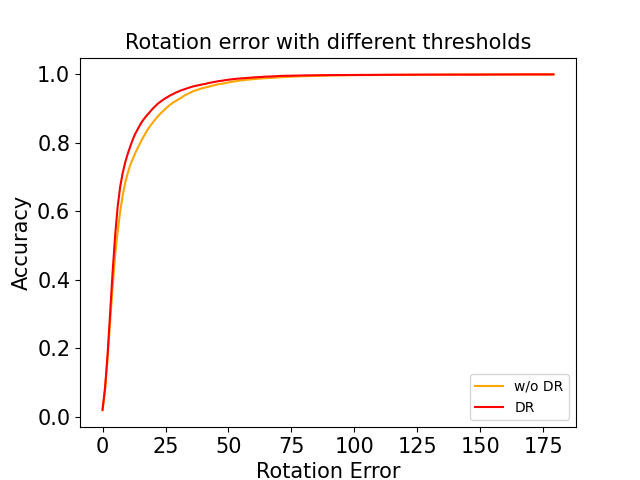}}\\
(a)Without Decoupled Training &(b)Decoupled Training&(c)Decoupled and w/o Decoupled
\end{tabular}
\caption{\textbf{Comparison on category-level}. The curve of accuracy under different rotation error thresholds with different training fashions on FS-Net.}
\label{fig:fsrrs}
\end{figure*}

\begin{figure*}[htp!]
\centering
\begin{tabular}{ccc}
{\includegraphics[width=0.3\linewidth]{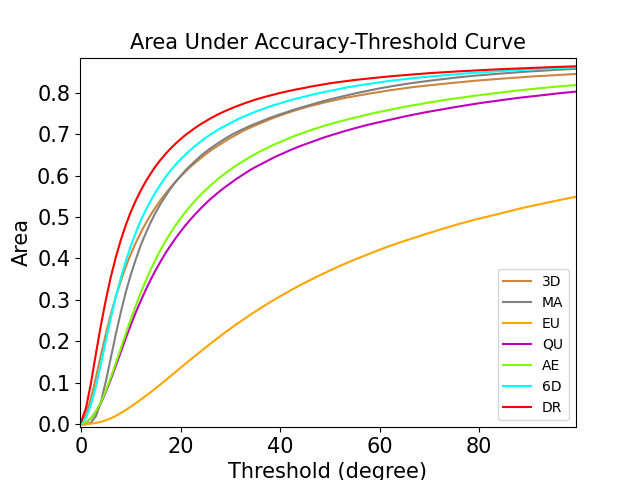}}&
{\includegraphics[width=0.3\linewidth]{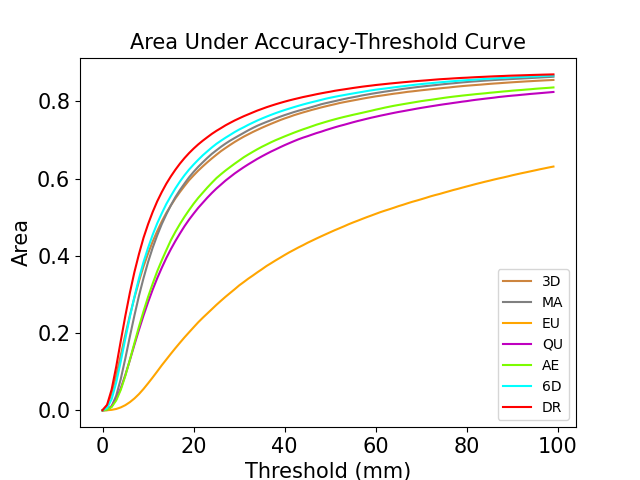}}&
{\includegraphics[width=0.3\linewidth]{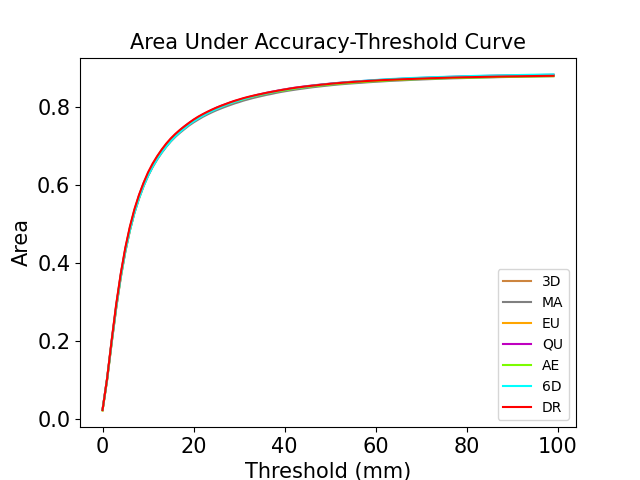}}\\
(a) AUC of Rotation &(b)AUC of ADD(S)  &(c) AUC of Translation Error
\end{tabular}
\caption{\textbf{Comparison on instance-level.} We report the AUC of rotation error, ADD(S) error and translation estimation error for different representation based on G2L-Net.}
\label{fig:g2lrrs}
\end{figure*}

\begin{table*}
\centering
\caption{\textbf{Rotation error comparison.}  For different representations on category-level 6D object pose estimation we provide the geodesic distance error of the rotation prediction on the test set. The `RM' means rotation mapping which is proposed in \cite{pitteri2019object}. `w/o decoupled' means training rotation as a whole term. 'decoupled' means training the rotation sub-terms with different network branches. `3D' means using the eight 3D corners to represent the rotation as in G2L\cite{Chen_2020_CVPR}. `$R_{6D}$' is the rotation representation proposed in \cite{zhou2019continuity}.}
\label{tab:fsrots}
\begin{threeparttable}

\begin{tabular}{clccccccc}
\hline
 &Category & Bottle & Bowl & Can & Camera & Laptop & Mug &  Average \\ \hline
\multirow{7}{*}{w/o decoupled}&FS-Net+3D+RM& 6.14 & 5.08 &\textbf{4.38} & 25.55 & \textbf{7.06} & 19.42 & 11.27\\
&FS-Net+Matrix+RM& 8.94 & \textbf{4.84} & 5.33 & 41.87 & 17.70 & 32.92 & 18.60\\
&FS-Net+Euler+RM& 13.06 & 6.18 & 7.23 & 44.50 & 13.92 & 37.39 & 20.38\\
&FS-Net+Quaternion+RM& 7.24 & 5.07 & 5.81 & 41.94 & 13.08 & 27.47 & 16.77\\
&FS-Net+Axis-angle+RM& 10.30 & 6.96 & 10.33 & 41.00 & 16.07 & 31.41 & 19.35\\
&FS-Net+$R_{6D}$+RM& 8.51 & 5.12 & 4.98 & \textbf{23.35} & 9.27 & \textbf{17.13} & 11.40\\
&FS-Net+FVR+RM& \textbf{5.88} & 4.92 & 4.40 & 24.72 & 7.96 & 19.12 & \textbf{11.17}\\
\hline
\multirow{4}{*}{decoupled}&FS-Net+Axis-angle+RM& 17.27 & 12.92 & 16.18 & 39.14 & 18.08 & 25.54 & 21.52\\
&FS-Net+Euler+RM& 17.46 & 16.91 & 19.05 & 52.58 & 40.30 & 55.81 & 33.68\\
&FS-Net+Quaternion+RM& 15.25 & 12.78 & 15.96 & 41.52 & 36.65 & 41.56 & 27.29\\

&\ronetwo{FS-Net+$R_{6D}$}& {5.07} & 4.36 & 5.17 & 23.91 & 8.74 & 13.98 & 10.21\\

&FS-Net+FVR \cite{Chen_2021_CVPR}& {5.71} & {3.85} & {3.97} & {21.59} & {8.00} & {13.77} & {9.48}\\
&\ronetwo{FS-Net+FVR+$\theta_g^*$+$\theta_r^*$+$L^*$}\tnote{$\Diamond$} & \textbf{4.98} & \textbf{3.43} & \textbf{3.18} & \textbf{20.39} & \textbf{7.14} & \textbf{11.02} & \textbf{8.36}\\
\hline
\multirow{2}{*}{\rone{different losses}}

&\rone{FS-Net+$R_{6D}$+smooth$_{L_1}$}& 5.12 & 4.21 & 4.50 & 20.45 & 9.12 & 18.84 & 10.37\\
&\rone{FS-Net+FVR+smooth$_{L_1}$}& 5.20 & 3.83 & 4.02 & 19.67 & 9.92 & 17.81 & 10.08\\

 \hline
\end{tabular}

  \begin{tablenotes} 
		\item[$\Diamond$] For symmetric objects `Bottle', `Bowl' and `Can', we only use $L^{*}$.
  \end{tablenotes}
\end{threeparttable} 
  
\end{table*}

\subsection{Rotation Representation Comparison}
In this section, we compare the FVR representation with other rotation representations on 6D object pose estimation. For the sake of fair comparison and following the experimental protocol in \cite{zhou2019continuity}, we adopt MSE loss for all rotation representations. \ronetwo{As described in Section \ref{sec:flex_fvr}, the FVR representation has four parameters.
For simplicity, we set $L_g=L_r=L$, then we find the optimal value of $L$ and $\theta$ separately. We first search the optimal value of $L$ by fixing the value of $\theta_r$ and $\theta_g$, then we search the optimal value of $\theta_g$ by fixing the value of $\theta_r$ and $L$. Finally, we search the optimal value of $\theta_g$ using the optimised $\theta_r$ and $L$.} 
\ronetwo{For optimization of $L$, we search the optimal value of $L$ in $\{1, 10, 50, 100, 200, 500, 1000\}$.}
\ronetwo{For optimization of the optimal $\theta_g$: by keeping $L$ and $\theta_r$ fixed, we gradually change the value of the $\theta_g$ from 0$^\circ$ to 360$^\circ$ with a step of 30$^\circ$.}
\ronetwo{For optimization of $\theta_r$: as in the above steps, we have found the optimal value of $L^{*}$ and $\theta_g^*$. We then set $L_g=L_r=L^{*}$ and $\theta_g=\theta_g^*$, and find the optimal value of $\theta_r$ by gradually changing the value of the $\theta_r$ from 0$^\circ$ to 360$^\circ$ with a step of 30$^\circ$.}

%

For category-level pose estimation, we use the proposed FS-Net pipeline as the baseline method. Same as \cite{zhou2019continuity} we use the geodesic errors to measure different rotation representations. We report the results on Table \ref{tab:fsrots} and Fig. \ref{fig:fsrrs}. 
For instance-level pose estimation, we use our previous work G2L-Net \cite{Chen_2020_CVPR} and the state-of-art method GDR-Net \footnote{\footnotesize{Code provided in \url{https://github.com/THU-DA-6D-Pose-Group/GDR-Net}}} \cite{Wang_2021_GDRN} as baseline to test different rotation representations. We summarise the results in Table \ref{tab:g2lrots} and Fig. \ref{fig:g2lrrs}.

In Table \ref{tab:fsrots}, in the `decoupled' part, we decouple the Axis-angle representation as two sub-terms: vector part and the length of the vector; we decouple Euler representation as three angles, we decouple Quaternion as four sub-terms which are four variables in the quaternion representation, and we decouple $R_{6D}$ as two vectors.

From Table \ref{tab:fsrots}, we can see that our new proposed FVR with optimised parameters can achieve the best performance for rotation estimation. 
Compared with the $R_{6D}$ rotation representation proposed by Zhou et al. \cite{zhou2019continuity}, in category-level pose estimation, our rotation prediction error is 18.11\% smaller than that of $R_{6D}$ representation (10.21 vs 8.36) in decoupled setting. When comparing FVR representation in different training fashions, the decomposable property of FVR can decrease the rotation error by 15.13\%, from 11.17 to 9.48. 
\ronetwo{To further prove that the proposed FVR is better than the $R_{6D}$, apart from MSE loss, we use the smooth$_{L_1}$ loss \cite{girshick2015fast} to train FVR and $R_{6D}$. From Table \ref{tab:fsrots}, we can see that with smooth$_{L_1}$ loss, our FVR representation can still achieve better results (lower rotation error) than $R_{6D}$ (10.08 vs 10.37). Then, we also train the $R_{6D}$ representation in the decoupled fashion. Table \ref{tab:fsrots} shows that under decoupled setting, our proposed FVR achieves lower rotation error than $R_{6D}$ (9.48 vs 10.21).}


\begin{table*}[t]
\caption{\textbf{Rotation error for different representations on instance-level 6D pose estimation.} The best result is bolded. `PM' means Point-Matching loss. }
\label{tab:g2lrots}
\centering
\resizebox{0.99\linewidth}{20mm}
{
\begin{tabular}{lcccccccccccccc}
\hline
Method &  Ape &  Bench Vise  & Camera & Can & Cat & Driller & Duck &Egg Box & Glue & Hole Puncher & Iron & Lamp & Phone& Average\\
\hline
G2L \cite{Chen_2020_CVPR}& {5.07} & 15.34 & 5.99 & 8.96 & {6.03} & 13.04 & 6.84 & 25.13 & 11.31 & 5.13 & 17.20 & 12.54 & 7.24 & 10.75\\
G2L+Matrix& 7.86 & 11.73 & 8.28 & 7.76 & 9.14 & 9.83 & 8.58 & {17.02} & 13.36 & 6.69 & 8.26 & 13.49 & 9.87 & 10.14\\
G2L+Euler& 42.79 & 54.67 & 42.33 & 53.59 & 43.07 & 46.06 & 39.41 & 52.32 & 49.91 & 51.66 & 45.03 & 53.87 & 48.20 & 47.92\\
G2L+Quaternion& 13.29 & 17.36 & 12.35 & 18.70 & 12.42 & 14.54 & 15.28 & 22.52 & 17.32 & 13.52 & 16.91 & 20.70 & 16.26 & 16.24\\
G2L+Axis-angle & 14.60 & 19.12 & 9.37 & 13.70 & 10.70 & 17.74 & 10.91 & 20.46 & 18.44 & 10.13 & 13.65 & 18.18 & 13.91 & 14.69\\
%
\rtwo{G2L+$R_{6D}$\cite{zhou2019continuity}}&6.77  & 11.64  & 6.15  & 8.23  & 5.97  & 10.08  & 7.16  & 21.35  & 14.17  & 5.16  & 7.78  & 10.76  & 6.78  & 9.38 $\pm$ 0.18\\
\rtwo{G2L+FVR}&6.31  & 12.21  & 5.50  & 6.73  & 6.71  & 11.29  & 8.11  & 16.29  & 13.18  & \textbf{3.55}  & 7.13  & \textbf{8.91}  & 5.48  &   8.57 $\pm$ 0.07\\

\ronetwo{G2L+FVR+$L^*$}& \textbf{4.77} & 13.02 & \textbf{4.32} & 6.00 & 5.03 & 14.64 & \textbf{4.91} & \textbf{15.33} & 12.76 & 4.20 & \textbf{4.87} & 9.46 & \textbf{5.20} & 8.04\\

\ronetwo{G2L+FVR+$\theta_g^*$}& 5.67 & \textbf{7.30} & 4.91 & 6.54 & 5.08 & 11.20 & 5.89 & 15.82 & 8.26 & 4.04 & 9.94 & 9.87 & 5.46 & 7.69\\

\ronetwo{G2L+FVR+$\theta_g^*$+$\theta_r^*$+$L^*$}& 5.15 & 11.68 & 6.17 & \textbf{5.87} & \textbf{3.76} & \textbf{7.45} & 5.06 & 15.62 & \textbf{9.55} & 4.33 & 7.62 & 10.68 & 5.40 & \textbf{7.56}\\
\hline
GDR-Net\cite{Wang_2021_GDRN}(+$R_{6D}$(PM))
& 2.11 & 1.85 &1.81 &1.82 &2.02&2.02&1.98&1.72&2.37 & 1.96 & 2.33 & 1.87 & 2.30 & 2.01\\


GDR-Net+$R_{6D}$ & 2.73 & 2.71 & 2.80 & 2.71 & 2.71 & 2.88 &2.77 & 2.49 & 3.18 & 2.98 &  3.04 & 2.74 & 3.43 & 2.86\\



GDR-Net+FVR & \textbf{1.83} & \textbf{1.82} &\textbf{1.67} &\textbf{1.73} &\textbf{1.80}&\textbf{1.83}&\textbf{1.83}&\textbf{1.58}&\textbf{1.95} & \textbf{1.88} & \textbf{2.04} & \textbf{1.68} & \textbf{2.19} & \textbf{1.83}\\

%
\hline
\end{tabular}
}
\end{table*}

From Table \ref{tab:g2lrots}, we find that the proposed FVR representation achieves the best performance for both G2L-Net and GDR-Net. \ronetwo{When using the searched optimal parameters, the rotation error can be further decreased from 8.57 to 7.56, a decrease of 11.8 \%.}
Compared with the $R_{6D}$ representation, in G2L-Net, FVR decreases the rotation error from 9.38 to 7.56, a decrease of 19.40\%. 
In GDR-Net, we retrain the network with $R_{6D}$(PM), $R_{6D}$ and FVR representation. All other parameters are fixed except the rotation loss part. Compared with $R_{6D}$(PM) and $R_{6D}$, our proposed FVR decreases the rotation error by 8.9\% and 36\%. 

In additional, we show the translation estimation error for different representations in Fig. \ref{fig:fsrrs}.c.
From Fig. \ref{fig:fsrrs}.c, we can see that the area under threshold curves of translation of different rotation representations are almost overlapped with each other, which means the rotation representation can merely affect the accuracy of rotation estimation.

\section{Conclusion}
In this paper, we propose a fast category-level pose estimation method that runs at 20 FPS, which has the potential for real-time applications. 
The proposed method extracts the latent feature by the observed points reconstruction with a 3DGC-based autoencoder. 
Then the category-level orientation feature is decoded by the proposed FVR representation. In addition, we also provide a simple search-based solution to find the optimal parameters of FVR representation. Though simple yet effective, this strategy may not fully explore the capacity of the proposed FVR, we believe the performance can be further boosted with more advanced optimization techniques.
For translation and object size estimation, we use the residual network to estimate them based on residual estimation. 
Finally, to increase the generalization ability of FS-Net and save the hardware source, we design an online 3D deformation mechanism for training set augmentation.
Extensive experimental results demonstrate that FS-Net is less data-dependent and can achieve state-of-the-art performance on category- and instance-level pose estimation in terms of both accuracy and speed. 
The experimental results also show that the proposed FVR representation is cable to find a more suitable vector-based rotation representation for 6D object pose estimation tasks than other widely used rotation representations. 

Please note that 3D augmentation and FVR representation are model-free, which can be easily plugged into other pose estimation methods to boost performance.  

\bibliographystyle{IEEEtran}
\bibliography{egbib}

%
%
\vspace{-1.5cm}
\end{document}